\documentclass[Afour,sageh,times]{sagej}

\usepackage{moreverb,url}

\usepackage[colorlinks,bookmarksopen,bookmarksnumbered,citecolor=red,urlcolor=red]{hyperref}

{}
{}
{}

\usepackage{acronym}
\usepackage{amsmath} 
\usepackage{mathrsfs}
\usepackage{bm}
\usepackage{amssymb}
\usepackage{amsfonts}
\usepackage{xfrac}
\usepackage{graphicx}
\usepackage[usenames,dvipsnames]{xcolor}
\usepackage{acronym}
\usepackage{float}
\usepackage{paralist}
\usepackage{enumitem}
\usepackage{epsfig}
\graphicspath{{Figures/}}
\usepackage{url}
\usepackage{stfloats}
\usepackage{amsopn}     % to get \DeclareMathOperator
\usepackage{psfrag}

\usepackage{mdwtab}     %ERW: added to get pretty tables
\usepackage{commath}
\usepackage{balance}
\usepackage{bm}
\usepackage{acronym}
\usepackage{balance}
\usepackage{tikz} 
\usetikzlibrary{arrows,calc,positioning,shadows,shapes}
\usetikzlibrary{fit}
\usepackage{adjustbox}
\usepackage{pgf} 
\usepackage{paralist}
\usepackage{mathtools, nccmath}
\usepackage{todonotes}
\usepackage[]{caption, subcaption}
\usepackage[normalem]{ulem}
\usepackage{comment}
\newcommand{\BP}{\mathbb{P}}

\newcommand{\filt}{\mathscr{F}}

\acrodef{snr}[\textsc{snr}]{signal-to-noise-ratio}
\acrodef{imu}[\textsc{imu}]{inertial measurement unit}
\acrodef{gps}[\textsc{gps}]{global positioning system}
\acrodef{sde}[\textsc{sde}]{stochastic differential equation}
\acrodef{ou}[\textsc{ou}]{Ornstein-Uhlenbeck}
\acrodef{dspp}[\textsc{dspp}]{doubly stochastic Poisson process}
\acrodef{smp}[\textsc{smp}]{stochastic maximum principle}
\acrodef{fbsdes}[\textsc{fbsde}s]{forward-backward stochastic differential equations}
\acrodef{pde}[\textsc{pde}]{partial differential equation}
\acrodef{cps}[\textsc{cps}]{counts per second}
\acrodef{mav}[\textsc{mav}]{micro aerial vehicle}
\acrodef{vio}[\textsc{vio}]{visual inertial odometry}
\acrodef{uas}[\textsc{uas}]{unmanned aerial system}
\acrodef{cots}[\textsc{cots}]{commercial off the shelf}
\acrodef{lrt}[\textsc{lrt}]{likelihood ratio test}
\acrodef{pmd}[\textsc{pmd}]{probability of missed detection}
\acrodef{pfa}[\textsc{pfa}]{probability of false alarm}
\acrodef{qp}[\textsc{qp}]{quadratic program}
\acrodef{gazebo}[\textsc{gazebo}]{}
\acrodef{rviz}[\textsc{rviz}]{}
\acrodef{2d}[\textsc{2d}]{}
\acrodef{3d}[\textsc{3d}]{}
\acrodef{ros}[\textsc{ros}]{}
\acrodef{gudhi}[\textsc{gudhi}]{}
\acrodef{lidar}[\textsc{l}i\textsc{dar}]{}
\acrodef{NLopt}[\textsc{NLopt}]{}
\acrodef{vicon}[\textsc{vicon}]{}
\acrodef{gurobi}[\textsc{gurobi}]{}
\acrodef{rgbd}[\textsc{rgb-d}]{}
\acrodef{sprt}[\textsc{sprt}]{sequential probability ratio test}
\acrodef{fov}[\textsc{fov}]{field of view}
\acrodef{esdf}[\textsc{esdf}]{Euclidean signed distance field}
\acrodef{fov}[\textsc{fov}]{field of view}
\acrodef{sfc}[\textsc{sfc}]{safe flight corridor}
\acrodef{vr}[\textsc{vr}]{Vietoris-Rips}
\acrodef{mpc}[\textsc{mpc}]{model predictive control}
\acrodef{kd}[\textsc{kd}]{kd}
\acrodef{cog}[\textsc{cog}]{cog}
\acrodef{msckf}[\textsc{msckf}]{msckf}
\acrodef{vi}[\textsc{vi}]{visual-inertial}
\acrodef{so}[\textsc{so(3)}]{so(3)}
\acrodef{cog}[\textsc{cog}]{center of gravity}
\acrodef{nuc}[\textsc{nuc}]{nuc}
\acrodef{cpu}[\textsc{cpu}]{cpu}
\acrodef{gpu}[\textsc{gpu}]{gpu}
\acrodef{nvidia}[\textsc{nvidia}]{nvidia}
\acrodef{ransac}[\textsc{ransac}]{ransac}
\acrodef{vins}[\textsc{vins}]{vins}
\acrodef{dji}[\textsc{dji}]{dji}
\acrodef{DM}[\textsc{dm}]{\textsc{d}ecision \textsc{m}aker}
\acrodef{slammot}[\textsc{slammot}]{simultaneous localization, mapping, and moving object tracking}
\acrodef{ssd}[\textsc{ssd}]{single shot detector}
\acrodef{yolo}[\textsc{yolo}]{you only look once}
\acrodef{iou}[\textsc{iou}]{intersection over union}
\acrodef{map}[\textsc{mAP}]{Mean Average Precision}
\acrodef{flann}[\textsc{flann}]{fast library for approximate nearest neighbors}
\acrodef{sift}[\textsc{sift}]{scale invariant feature transform}
\acrodef{dnn}[\textsc{dnn}]{deep neural network}
\acrodef{cnn}[\textsc{cnn}]{convolution neural network}
\acrodef{gm}[\textsc{gm}]{Gieger-Muller}
\acrodef{ram}[\textsc{ram}]{}

\newtheorem{problem}{Problem}

\DeclareMathOperator\erf{erf}

\definecolor{yaleblue}{rgb}{0.06, 0.3, 0.57}

\newcommand\BibTeX{{\rmfamily B\kern-.05em \textsc{i\kern-.025em b}\kern-.08em
T\kern-.1667em\lower.7ex\hbox{E}\kern-.125emX}}

\def\volumeyear{2016}

\begin{document}

\runninghead{Yadav, Sebok and Tanner}

\title{Receding Horizon Navigation and Target Tracking for Aerial Detection of Transient Radioactivity}

\author{Indrajeet Yadav\affilnum{1}, Michael Sebok\affilnum{1} and Herbert G Tanner\affilnum{1}}

\affiliation{\affilnum{1}University of Delaware, USA}

\corrauth{Indrajeet Yadav, University of Delaware,
126 Spencer Lab,
Newark,
DE-19716, USA.}

\email{indragt.yadav@udel.edu}

\begin{abstract}
The paper presents a receding horizon planning and control strategy for quadrotor-type \ac{mav}s to navigate reactively and intercept a moving target in a cluttered unknown and dynamic environment. 
Leveraging a lightweight short-range sensor that generates a point-cloud within a relatively narrow and short \ac{fov}, and an \acs{ssd}-MobileNet based Deep neural network running on board the \ac{mav}, the proposed motion planning and control strategy produces safe and dynamically feasible \ac{mav} trajectories within the sensor \acs{fov}, which the vehicle uses to autonomously navigate, pursue, and intercept its moving target. 
This task is completed without reliance on a global planner or prior information about the environment or the moving target. 
The effectiveness of the reported planner is demonstrated numerically and experimentally in cluttered indoor and  outdoor environments featuring maximum speeds of up to 4.5--5~m/s.
\end{abstract}

\keywords{Receding Horizon Motion Planning, Target Tracking, Reactive Obstacle Avoidance, Aerial Radiation Detection }

\maketitle

\section{Introduction}

\subsection{Overview}

The work reported in this paper is motivated in part by application problems in the field of nuclear nonproliferation.
In this context, there can be instances where one needs to quickly detect weak radiation sources that could be in transit through generally unknown, and sometimes \acs{gps}-denied environments.
With \acp{mav} being now a widely available and accessible technology, and with the possibility of mounting lightweight (albeit still of low efficiency) \ac{cots} radiation detectors on them, such a detection task is now possible using aerial means.

The approach to such problems advocated here utilizes an algorithmic pipeline that combines reactive receding horizon navigation with target tracking for \acp{mav}, a \ac{vi} state estimation algorithm, a \acs{ssd}-MobileNetV2 based visual target identification and tracking, and a fast likelihood ratio-based binary decision making algorithm.
%Without relying on grid-like environment representations, which are known to scale poorly with workspace size and resolution, the approach integrates an \acs{ssd}-MobileNetV2 based target tracking strategy to the motion planning pipeline to obtain real-time \acs{3d} target position information.
Reactive planning is achieved through a new \ac{mpc}-type motion planner that fully incorporates the nonlinear vehicle dynamics of the \ac{mav} (cf.~\citet{TAC10}) and utilizes real-time sensor data in the form of a point-cloud generated by an onboard \acs{rgbd} camera, to select a (probabilistically optimal) safe path toward the target within the field of view.
The reactive planner solves a multi-objective optimal control problem in real time, and thus balances detection accuracy and decision-making speed against platform flight agility limitations.
The product of this optimization is a dynamically compatible minimum snap trajectory that fits along the kinematic reference path. 
Then a nonlinear geometric controller on the \ac{mav} tracks this trajectory in a receding horizon fashion.

\begin{figure}[t!]
	\centering
	\includegraphics[width=1.0\linewidth]{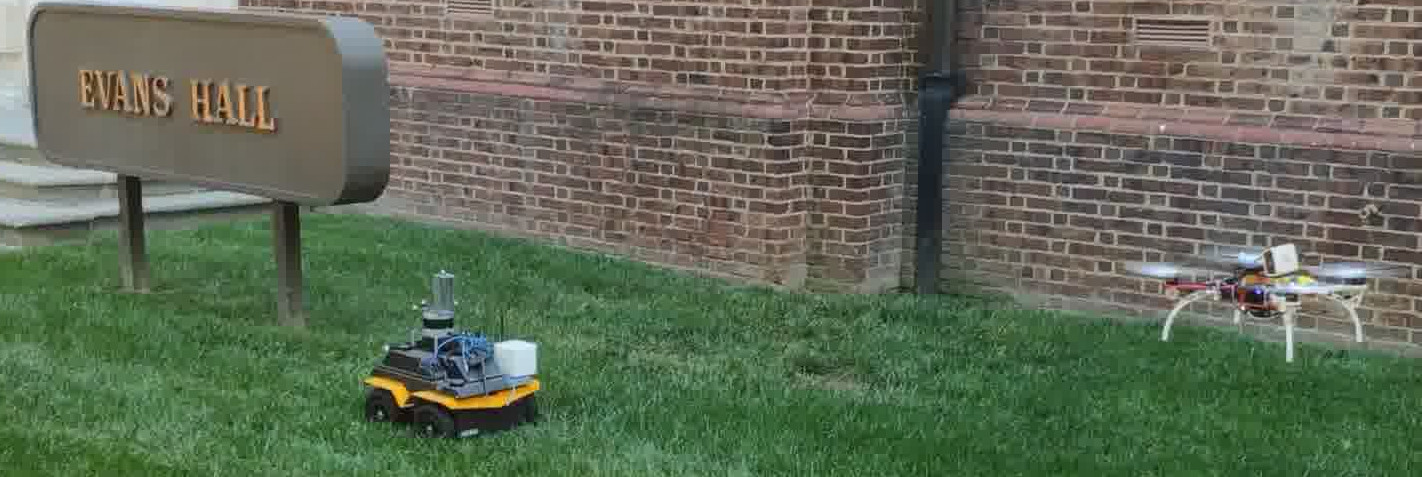}
	\caption{Custom-built quadrotor capable of on-board planning, control, state estimation, ssd-mobilenet based target tracking motion planning and radiation measurement tracking the Jackal (ground-vehicle).		\label{Fig1}
	}
	%\vspace{-0.5cm}
\end{figure}

The planning and control architecture is implemented using low-cost \ac{cots} hardware, limited-range sensors, and computation devices with average  capabilities. 
With these resources, the system has demonstrated speeds up to 4.5~m/s in cluttered indoor as well as outdoor environments. 
In the experimental results reported in this paper, the planner enables the \ac{mav} to intercept to static or dynamic ground targets.
Such interception maneuvers have been shown to increase the likelihood of correct identification of the presence of a radioactive source on the target~\citep{AURO-Jianxin}.
%under the assumption that the latter remain within the extension of the vehicle's \ac{fov}. 
In cases where the local information is insufficient to construct a feasible trajectory to the goal, the planner is designed to safely stop the vehicle.
The features and capabilities of the reported reactive planning and control architecture is demonstrated in case studies of mobile radiation detection, in which \acp{mav} autonomously determine whether radiation sources are present on a ground vehicle moving in cluttered environments (Fig.~\ref{Fig1}).

\subsection{Background and Related Work}
A quadrotor-type \ac{mav} is an inexpensive, lightweight, and agile sensor platform, suitable for many applications in areas including surveillance, aerial photography and mapping, precision agriculture, construction and defense. 
Although \acp{mav} with various degrees of autonomy have been deployed in these application domains, the prevailing assumption is that the environment is at least partially known, so that a motion plan can be generated a priori and then used for steering the vehicle to a desired goal.
To plan its motion in unknown environments, however, a robotic vehicle needs to construct and update a (local) environment map online; recent \ac{mav} literature addresses this problem using an onboard perception stack~\citep{voxblox,Hornung2013}.

\subsubsection{Perception-Based Reactive Navigation}

As long as vehicle dynamic constraints are respected, safe \ac{mav} navigation and aggressive maneuvering is possible by combining a reference trajectory generation process that splices together time polynomials  between waypoint poses (keyframes)~\citep{Mellinger,Richter2016}; 
a differential-geometric quadrotor point stabilization or tracking controller~\citep{Lee} then is utilized to transition between these keyframes.
While platform-specific constraints on vehicle dynamics can be identified by experimental testing, \emph{ensuring} safety during operation requires either complete prior knowledge of the environment, or some way of acquiring the missing information in real time through onboard sensing; both options involve  several nontrivial and open research questions~\citep{Cadena}.

Early work on online obstacle avoidance focused on  building a new, or updating a prior, map of the environment. 
In this context, a continuous-time trajectory optimization using {\small \textsf{octomap}}~\citep{Hornung2013} is employed~\citep{Oleynikova}, utilizing a local planner to re-generate a safe trajectory that assigns collision costs based on a (computationally expensive) \ac{esdf} map. 
In unknown environments, \cite{gao} construct a navigation approach utilizing online planning to produce a point-cloud map of the environment using a Velodyne \acs{3d} \acs{lidar}, and use the map to find safe corridors through which their \acs{mav} plans it motion. 
To navigate to destinations outside the sensor range, a sequence of predefined waypoints is needed.
\cite{landry} assume knowledge of obstacle location and geometry, and locally decompose the available free space into convex polyhedra to generate safe paths for the vehicle.
Variants of such approaches~\citep{sliu} consider the workspace represented as a \acs{3d} grid map with uniform voxels, which are used to create a convex \ac{sfc}.
%The former class of methods involves a global planner, while the latter ensures completeness only in the sense of finding an initial path through the given global map.   

More recent reactive motion planning algorithms specifically for aerial vehicles include lightweight egospace-based algorithms extended to a quadrotor's configuration dynamics~\citep{Fragoso}, or reactively sample safe trajectories in the field of view of the quadrotor, and decouple local obstacle avoidance from global guidance using a global planner~\citep{MRyll}. 
\citet{Sikang2016} report on an impressive \emph{receding horizon}-based approach to (local) planning that involves only limited onboard sensing, and which utilizes a local uniform resolution volumetric occupancy grid map and a cost map to find and navigate to safe frontier points (local goals that are closest to the global goal). 

All aforementioned approaches either require some type of prior information and a global planner that generates a sequence of waypoints,  or rely on (\ac{mav} payload-taxing) high-range sensors. 
%Guarantees of convergence to the desired destination are elusive here.
They are particularly effective for navigating to static configurations;  if, however, the \ac{mav} destination changes over time, or if the generation of a prior map is impossible, they cannot guarantee a priori the existence of a \ac{sfc}, or ensure that intermediate waypoints will be within sensor range.

Yet even within a reactive navigation framework, some convergence guarantees can still be provided~\citep{CDC21a}
based on appropriate extensions of the navigation function approach~\citep{Rimon}.
Recent advances in such techniques have independently demonstrated success in online calculation of convergent vector fields using limited range onboard sensors~\citep{Vasilopoulos_2020, Arslan}, where unknown obstacles are detected online using a \ac{dnn} trained on the predefined class of object geometries.
Such approaches have been successfully tested on fully actuated or differential drive robots at relatively low speeds; their applicability to fast moving \acp{mav} with underactuated higher order dynamics remains to be demonstrated.

%While impressive, this approach also considers static goals and its applicability to the dynamic goals is not clear.
%Another available high speed receding horizon planner works by sampling  safe trajectories as the quadrotor moves towards its goal~\cite{MRyll}. 
%Comparative studies and details can be found in Section~\ref{results}.

\subsubsection{Target Tracking}

While the area of \ac{slammot} has been the focus of recent research efforts~\citep{Wang2007IJRR, Chojnacki2018}, there are few implementations involving 
%visual-inertial sensors.
%One active tracking approach~\cite{ZhouKe2011TRO} explores the idea of planning the robot's motion in ways that minimize future target position uncertainty. 
%This methodology is developed for 2D problems and assumes a perfectly known sensor state. 
%Another 
active visual-inertial target tracking approach~\citep{Chen2016IROS} that have demonstrated impressive performance using a quadrotor. 
%yet, the size of tracking errors reported can be detrimental to detection of low-intensity mobile sources of radioactivity, because of the critical role that the distance between sensor and source plays in the context of nuclear measurement~\cite{Nemzek,PSPT_Automatica}.
Conceptually closer to the one reported in this paper, is the aerial target tracking work of \citet{VK1} who demonstrate tracking of a spherical rolling target.
That approach employed a geometric technique similar to visual servoing for tracking the target combined with a receding horizon strategy that penalizes velocity and position errors.
%, and a \ac{mav} motion control scheme based on minimum-snap trajectory generation~\cite{Mellinger}. 
Alternatively, if the environment is known, a reactive motion planning scheme can utilize multi-objective optimization for obstacle avoidance and target tracking~\citep{Penin}.

These approaches assume that the target is either known, detected via some type of a tag, or it can be localized using some form of visual servoing (which depends on the shape of the target). 
Alternative means of target detection include approaches based on \acp{cnn}, which have gained popularity in active quadrotor motion planning and target tracking with the advent of computationally efficient networks such as \ac{ssd}~\citep{Liu_2016},  \ac{yolo}~\citep{Redmon}, and their variants, and seem particularly effective at high sensor speeds.
For instance, DroNet architectures~\citep{Loquercio} are utilized in Drone Racing involving dynamically changing environments~\citep{Loquercio_2020}.  
In another related application of forest trail navigation, a multi-column \acs{dnn} is used to predict the turns in the trail and guide the quadrotor~\citep{trailfollow_Giusti}, although
that particular perception and control stack runs remotely on a laptop. 
\citet{smolyanskiy2017lowflying} utilize a \ac{yolo} network but runs the planning and perception stack onboard the quadrotor on an \acs{nvidia}~\textsc{tx1}. 
A human gaze-driven quadrotor navigation utilizing a \acs{ssd} network running on an \acs{nvidia}~\textsc{tx2}  features eye-tracking glasses along with a camera and an \ac{imu}, which are combined to estimate the position of the human with respect to the quadrotor and steer the \ac{mav}~\citep{Yuan}.     

As far as the state of the art in experimental implementations is concerned, recent literature reports quadrotor navigation results at impressively high speeds~\citep{sliu,MohtaK,MRyll}; yet most of the systems either  involve a global planner~\citep{MRyll}, or employ high-end and expensive sensors with extended range---e.g., Velodyne \textsc{vlp-16} or Hokuyo \textsc{ust-20lx} \textsc{l}i\textsc{idar} mounted on a gimbal to provide 270$^\circ$ \ac{fov}~\cite{sliu,MohtaK}, which can offer significantly more advanced perception capabilities compared to a 69.4$^{\circ}\times\;$42.5$^{\circ}$ \ac{fov} sensor utilized in this work. 
In addition, the top speeds reported by \citet{MRyll} were achieved in open areas and with \emph{average} flight speeds of 2.4~m/s.
In the absence of a global planner, however, a planner with myopic vision cannot arbitrarily increase the speed in anticipation of an unseen obstacle. 
%Higher speeds may be possible, but in the absence of a global planner the planner's frequency will inevitably be limited by the that of the onboard processor.
Moreover, the majority of experimental results reported relate to the case of converging to static final configurations.

%What is more, most of application scenarios considered in the recent 
Thus existing robotics literature on \acp{mav} covers either safe and agile point-to-point navigation in unknown environments, or autonomously tracking moving target, but there is scarce, if any, work on the combined problem. 
In addition, the signal and data processing component, that is, what will be done with the measurement data once they have been collected, is typically an afterthought; not much effort is directed to designing the measurement process so that it serves the needs of decision-making based on these data.

\subsection{Paper Organization and Contributions}

At a high level, the approach of this paper is different: motion planning, safe navigation, target tracking and decision-making components are integrated and co-designed to operate in concert with each other.
In this regard, this paper makes technical contributions to \emph{real-time, sensor-based reactive navigation and target tracking} in the form of a motion planning and control methodology applicable to problem instances involving either static or moving navigation goals.
The planning and control architecture is unique because:
\begin{enumerate}
\item it is applicable to cases with both static and moving goal configurations;
\item it generates trajectories in a new way by efficiently solving a multi-objective optimal control problem;
\item it is experimentally tested indoors and outdoors with \acs{mav} speeds of up to 4.5--5 m/s; and
\item it is demonstrated to be capable of solving a challenging radiation detection problem.
\end{enumerate}

%Beyond this section, the paper is structured as follows. 
%Section \ref{overview} describes the overall architecture of the methodology.  
%A more detailed description of radiation detection problem, \acs{mav} dynamics and control, planning, and target tracking components, is given in Sections \ref{nuclear detection}, \ref{dynamicsandcontrol}, \ref{section:RHP} and \ref{targettracking} respectively. 
%Implementation and testing results are discussed in  Section~\ref{results}, and the paper concludes with Section~\ref{conclusions}.

\section{Overview of the Architecture}\label{overview}
%\vspace{-0.1cm}

\begin{figure}[t!]
	\centering
	\includegraphics[width=1.0\linewidth]{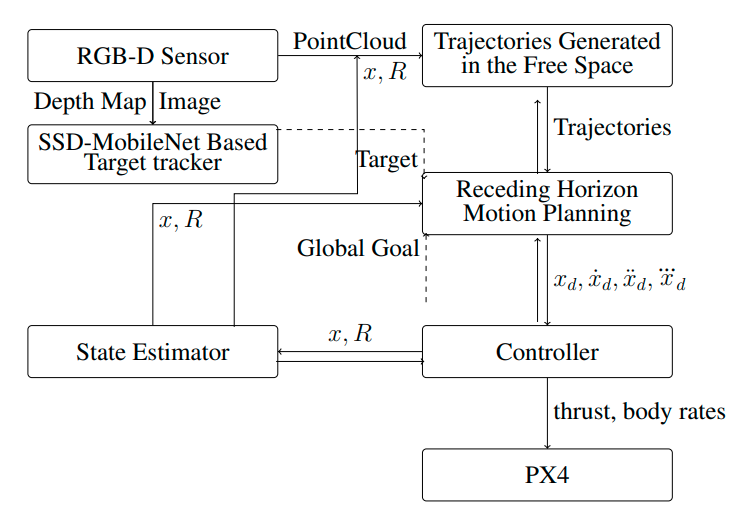}
	\caption{Block diagram of the   motion planning and control architecture.The arrows indicate the direction of information flow. The dashed lines indicate that the input to the planner is either a user specified static goal or relative position of the target obtained from \ac{ssd}-MobileNet.	\label{fig:sys_block_diag}
	}
	%\vspace{-0.5cm}
\end{figure}

The block diagram of Fig.~\ref{fig:sys_block_diag} shows the flow of information within the motion planning, state estimation, and platform control architecture. 
Starting with the \acs{rgbd} camera, a point-cloud is produced and then used to frame the obstacle-free portion of the workspace. % (Fig.~\ref{Fig6}).
%{\color{blue} as a set free points within the field of view of the sensor to which the robot could traverse from its current location (see representative Fig.~\ref{Fig3})}.
On board state estimation is performed through a visual-inertial \acs{msckf} navigation stack (Open-\acs{vins}~\citep{Geneva2020ICRA}). 
Given the free workspace and the state estimates, the receding horizon planner 
\begin{inparaenum}[(i)]
\item generates a candidate minimum snap trajectory from the current position of the robot to each of the free points in the field of view based on Pontryagin's minimum  principle;
\item selects the point in the \ac{fov} closest to the target (henceforth referred to as the \emph{intermediate point});  
\item assigns a cost to each trajectory as a weighted sum of the trajectory's proximity to obstacles and the proximity of the trajectory's end point to the intermediate point; and finally, 
\item selects the trajectory with the least such cost as the \emph{reference trajectory} for the \ac{mav} to follow. 
%The end point of the reference trajectory is referred to as the local goal in the reminder of the paper.
\end{inparaenum}
The end point of this selected minimum-snap reference trajectory is referred to as the \emph{local goal}.
An initial segment of the reference  trajectory that ends at the local goal is presented to a differential-geometric tracking controller, which initiates steering of the \ac{mav} to the designated local goal point within the \acs{fov}. 
In a typical receding horizon fashion, before the end of that initial segment is reached, the \ac{mav} uses an updated set of point-cloud data to generate a new reference trajectory, starting from a point on the current trajectory associated with the end of some designated control time horizon. 
Once in the neighborhood of this point, the \ac{mav} transitions smoothly to tracking of the new reference trajectory.
This constitutes a \emph{planning cycle} of duration $\delta t$.
The planning, and concurrent trajectory tracking, cycle repeats until the final  destination is reached ---when the latter is static.

If the destination is a moving target, an \ac{ssd} MobileNet-\textsc{v2} based target tracker calculates the relative positions between the target and the \acs{mav}, which the planner utilizes to generate intercepting trajectories.

\section{Radiation Detection Preliminaries}
\label{sec:nuclear detection}
\subsection{Application Case Study Background}
The robotics application that motivated this work was detecting weak mobile sources of radiation using  aerial mobile sensors. 
One possibility for realizing an inexpensive fully autonomous aerial radiation detector is to utilize a quadrotor \ac{mav}, mount a \ac{cots} radiation counter, and allow it to maneuver itself in ways that allow it to determine as quickly as possible whether or not its given moving target is radioactive or not. 
%Radiation detection algorithms assume that the arrival statistics of gamma-rays incident on a detector is a Poisson process~\cite{snyder}.
%In our case, since the relative distance between sensor and source is time-varying, this Poisson process is actually \emph{inhomogeneous}.
While a network of static detectors can also be deployed to the detection, due to the strong (inverse square law) dependence of the process intensity on the distance between sensor and source~\citep{NDTW}, the static network will experience a dramatic decrease of its \ac{snr} that will significantly reduce its detection accuracy and speed.

%In that vein, threat-based coverage over a given search area can be enhanced by allowing limited sensor mobility~\cite{CYTMA}. 
%Mobile sensors can also be steered through an information gain-driven search~\cite{Ristic}, although such approaches work well mainly for high intensity radioactive sources. 
%Wherever mobility is utilized, however, the question of how exactly sensor mobility---and position uncertainty, in particular---affects detection accuracy and speed is not directly addressed.
Detecting a mobile radioactive source based on a series of discrete events associated with the arrival of particles or gamma rays (indistinguishable from each other), can be formulated as an inhomogeneous \ac{lrt} within the Neyman-Pearson framework~\citep{PSPT_Automatica}. 
In this framework, \emph{analytical} (Chernoff) bounds on error probabilities can be derived in terms of relative sensor-source distance and used as proxies for the otherwise intractable true detection error probabilities~\citep{PSPT_Automatica}, to facilitate the design of optimal control laws that maximize detection performance in terms of speed and accuracy~\citep{AURO-Jianxin,Yadav}. 
%The practical and intuitive insights of the aforementioned work is that sensors should always be driven close to the potential source as quickly as possible.

%\vspace{-0.3cm}
\subsection{Neyman-Pearson based Radiation Detection}

A sensor measurement event---for the case of a \ac{gm} detector, for instance---is associated with the discrete random process where a gamma-ray (from the source or background) hits the sensitive area of the detector and records a ``count.''
This is mathematically modeled in a probabilistic setup  \citep{PSPT_Automatica,AURO-Jianxin} as follows.  
Assume that there is a counting process $N_t$ for $t \in [0,T]$, on a measurable space $(\Omega,\filt)$. 
In this context, $N_t$ essentially represents the number of times gamma-rays have hit the radiation counter located at position $x\in \mathbb{R}^3$, up to (and including) time $t \in [0,T]$. 
This counting process is assumed to be a Poisson process~\citep{snyder}. 
The sensor position $x$ is known over the whole interval $[0,T]$.

The detector can make two competing hypotheses, $H_0$ and $H_1$, each expressing an opinion as to whether its cumulative count rate can be attributed to the presence of a radiation source of intensity $a$ (in counts per unit time, i.e., \ac{cps}), located at a possibly time-varying position $y \in \mathbb{R}^3$ which will be referred to as the \emph{target}.
Hypothesis $H_0$ asserts that the target is benign, while hypothesis $H_1$ claims that the target carries a source of intensity.
The two hypotheses $H_0$ and $H_1$ correspond, respectively, to two distinct probability measures $\BP_0$ and $\BP_1$ on $(\Omega,\filt)$. 
With respect to measure $\BP_0$, the process $N_t$ is a Poisson process with intensity $\beta(t)$, i.e., the intensity of naturally occurring background radiation; with respect to $\BP_1$, however, the same process is Poisson with intensity $\beta(t)+\nu(t)$, where $\nu(t)$ represents the intensity of the source (whenever present) as perceived by the sensor at time $t$. 
The functions $\beta(t)$ and $\nu(t)$ defined on $[0,T]$ are assumed to be bounded, continuous and strictly positive~\citep{PSPT_Automatica}. 

Function $\nu(t)$ should implicitly reflect the inverse square law dependence of the source intensity perceived by the sensor on the distance between the sensor and the source~\citep{NDTW}. 
%The functional representation used here encodes the inverse square law~\cite{NDTW} relationship between range and perceived intensity, and the latter can be time varying because the distance between the two points can change over time.  
If $\chi$ denotes the sensor's cross-section coefficient, one possibility for expressing the detector's perceived intensity is in the form:
\begin{equation}\label{Eq:perceived_intensity}
\nu(t) = \frac{\chi a}{2\chi + \| y(t) - x(t) \|^2}  
\enspace.
\end{equation}

A test for choosing between $H_0$ and $H_1$ faces the risk of two types of errors. 
One of them is a \emph{false alarm}, which occurs when the outcome of the test is in favor of $H_1$ while $H_0$ is instead the correct hypothesis; the other is a \emph{missed detection} in which one sides with $H_0$ when in fact $H_1$ is true. 
%A test for choosing between $H_0$ and $H_1$ is considered an event $B_1$.
%The occurrence of event $B_1$ is ascertained on the basis of sensor observations over $[0,T]$, and has the following significance: for an outcome $\omega$, if $\omega \in B_1$, decide $H_1$; otherwise, that is if $\omega \in B_0 \triangleq \Omega\setminus B_1$, decide $H_0$. 
%For such a test, two types of errors can occur. 
%A \emph{false alarm} occurs if $\omega \in B_1$ with $H_0$ being the correct hypothesis; this occurs with probability $\BP_0(B_1)$. 
%A \emph{missed detection} occurs if  $\omega \in B_0$ while $H_1$ is the true hypothesis; this has probability $\BP_1(\Omega \setminus B_1)$. 
In this setting, the optimal test for deciding between $H_0$ and $H_1$ is an \ac{lrt} obtained as follows \citep{PSPT_Automatica}.
Let $\tau_n$ for $n \ge 1$ denote the $n$\textsuperscript{th} jump time of $N_t$ (jumps occur when the sensor registers a count), and with the convention that $\prod_{n=1}^0 (\cdot)=1$, let
%\footnote{$L_T$ is the likelihood ratio \cite{vantrees}.}
\begin{align}\label{Eq:LT}
L_T  =  \exp\left(-\int_{0}^{T} \nu (s) \dif s\right) \prod_{n=1}^{N_t}  1+ \frac{\nu (\tau_n)}{\beta (\tau_n)}
\enspace.
\end{align}
%Assume that $P_1$ is absolutely continuous with $P_0$, and that $H_0$ and $H_1$ are equiprobable~\cite{snyder}.
be the \emph{likelihood ratio}. 
Then for a specific fixed \emph{threshold} $\gamma>0$, the test
\begin{align}\label{Eq:LRT}
 L_T \mathop{\gtrless}_{H_0}^{H_1} \gamma  
\end{align}
is optimal in the (Neyman-Pearson) sense.
This means that \eqref{Eq:LRT} minimizes the probability of a missed detection under a given upper bound constraint on the probability of false alarm.
%, meaning that if $A_2$ is any other test whose probability of false alarm $P_0(A_2) \leq P_0(L_T \geq \gamma)$, then the probability of a miss(ed detection) for \eqref{Eq:LRT} is at least as low as that for $A_2$, i.e.\ $P_1(L_T < \gamma) \leq P_1(\Omega \setminus A_2)$. 
With $\mu(t) \triangleq 1+ \tfrac{\nu(t)}{\beta(t)}$, constants $p \in (0,1)$ and $\eta \triangleq \log \gamma$, and the quantity
\begin{equation}\label{Lambda}
\Lambda(p) \triangleq  \int_{0}^{T} \left[\,\mu(s)^p- p \,\mu(s) +p -1\right] \beta(s) \dif s 	\enspace,
\end{equation}
one can now extract analytical expressions for Chernoff bounds on the probabilities of false alarm $P_F$ and missed detection $P_M$~\citep{PSPT_Automatica}. 

%one can express Chernoff bounds on the probability of false alarm $P_F$ and missed detection $P_M$ as~\cite{PSPT_Automatica}

%\begin{align*}%\label{Eq:5}
%P_F & \leq \exp \left(\, {\textstyle \inf_{p>0}} %[\Lambda(p)-p\,\eta\,]\,\right) \\
%P_M & \leq \exp \left( {\textstyle \inf_{p<1}} %[\Lambda(p)+(1-p)\eta]\right)
%\enspace.
%\end{align*}

If an upper limit $\alpha>0$ is set on the bound on probability of false alarm, then there exists a unique solution $p^{\ast} \in [0,1]$ 
%to $\exp \big( {\textstyle \inf_{p<1}} [\Lambda(p)+(1-p)\eta]\big) = \alpha$ 
for which the tightest bound on the probability of missed detection can be obtained. 
The exponent in the bound on the probability of false alarm and missed detection, respectively, is~\citep{PSPT_Automatica} 
\begin{align}\label{Eq:PF}
&\mathcal{E_F} =  \int_{0}^{T} 
	[p^{\ast}\mu^{p^{\ast}} \log \mu - \mu^{p^{\ast}}+1] 
    \beta \dif s = -\log \alpha		\\
&\mathcal{E_M} = \log \alpha + \Lambda^{'}(p^{\ast}) \label{Eq:PM} 
\enspace,
\end{align}
where the dependency of $\mu$ and $\beta$ on time is suppressed for clarity, and derivative $\Lambda'(p) = \pd{\Lambda}{p}$ is expressed as%~(cf.~\cite[\S2.4]{vantrees}) 
\begin{equation} \label{LambdaPrime}
\Lambda^{'}(p)=  \int_{0}^{T} [\mu^p \log \mu - \mu+1] \beta \dif s
\enspace.
\end{equation}

For the optimal $p^{*}$, the $\Lambda(p^{*})$ and detection threshold $\gamma$ are related in the form $\gamma = \exp\big(\Lambda(p^{*})\big)$. 
Suppose now that the distance between target and sensor, $\|y-x\|$, is regulated by a control input ${u}$; then $\nu$, and consequently $\mu$, depend implicitly on ${u}$ and an optimal control problem can be formulated as follows:
\begin{problem}
Find a control input ${u}$ that optimizes $\Lambda^{'}(p^{\ast})$ for a given upper limit $\alpha$ on the bound on the probability of false alarm. 
\end{problem}
Irrespective of whether $\|y-x\|$ is deterministic or stochastic, it can be shown that the optimal sensor management strategy ${u}$ for sensors is to close the distance between source and sensor as quickly as possible~\citep{AURO-Jianxin, IY}. 
%The case where $\|y-x\|$ is  random  has been an open problem.

\section{Quadrotor Dynamics and Control}\label{dynamicsandcontrol}

The \ac{mav} is modeled as a rigid body moving in $\mathsf{SE}(3)$. 
Let $m$ and $\mathbf{J}$ denote its mass and moment of inertia, respectively, and denote $\mathsf{x} = (x,y,z)^\intercal$ and  $\mathsf{v}=\dot{\mathsf{x}}$ its 3\textsc{d} position and velocity in the inertial frame. 
Let $\mathbf{R}$ be the rotation matrix from the body-fixed frame to the inertial frame, and $\Omega$ be the \ac{mav}'s angular velocity in the body-fixed frame. 
We denote $\hat{\cdot}$ the skew symmetry operation, and write the gravitational acceleration vector and the unit vector along the inertial vertical as $\mathsf{g} = (0, 0, g)^\intercal$, and  
$\mathsf{e}_3 = (0,0,1)^\intercal$, respectively.
The control inputs to the \ac{mav} are the (scalar) thrust $f$ and the (vector) moment $\mathsf{M}$, both of which are considered bounded.
The former is bounded in the interval $(f_\mathrm{min},f_\mathrm{max})$;
the constraint on the latter, is assumed to translate to a bound on angular velocity $\|\Omega\| \leq \Omega_\mathrm{max}$.
With these definitions and assumptions in place, the \ac{mav} dynamics is described as%\useshortskip
\begin{subequations}
\label{uav-dynamics}
\begin{flalign}
 \dot{\mathsf{x}} = \mathsf{v},  
\qquad m\;  \dot{\mathsf{v}} =  f\; \mathbf{R}\; \mathsf{e}_3 -m\; \mathsf{g} 
\\
 \dot{\mathbf{R}} = \mathbf{R} \;  \hat{\Omega},  
\qquad \mathbf{J}\; \dot{\Omega} + \Omega \times  \mathbf{J} \;\Omega = \mathsf{M} 
\enspace.
\end{flalign}
\end{subequations}

The control law for this system is designed based on a (differential) geometric method~\citep{Lee}.
%with minor modifications. 
To examine the particular control design employed here in more detail, consider a smooth position reference trajectory $\mathsf{x}_d(t) \in \mathbb{R}^3$.
Based on this one can construct~\citep{Lee} a desired rotation matrix $\mathbf{R_d}$ and angular velocity $\Omega_d$ that are consistent with the reference trajectory $\mathsf{x}_d(t)$.
%This paper slightly departs from the design of \citet{Lee} in the sense that a desired 
The desired yaw angle $\psi$ of the \ac{mav} is used to construct the desired \ac{mav} direction vector in the form $\mathsf{b}_{1d}=(\cos{\psi}, \sin{\psi}, 0)^\intercal$.

The tracking errors in position $\mathsf{e_x}$, velocity $\mathsf{e_v}$, orientation $\mathsf{e_R}$, and angular rate $\mathsf{e_\Omega}$, are expressed as
\begin{subequations}
\label{errors}
\begin{align}
\mathsf{e_x} &= \mathsf{x}-\mathsf{x}_d
&
\mathsf{e_v} &=\mathsf{\dot{x}}-\mathsf{\dot{x}}_d\\
\mathsf{e}_R &= \frac{1}{2}(\mathbf{R}^\intercal_d \mathbf{R} - \mathbf{R^\intercal}\mathbf{R}_d)
&
\mathsf{e_\Omega} &= \Omega - \mathbf{R^\intercal}\mathbf{R}_d \Omega_d 
\enspace.
\end{align}
\end{subequations}

Picking positive control gains $k_x$, $k_v$, $k_R$ and $k_\Omega$, the control inputs can now be constructed as
\begin{subequations}
\label{mav-control}
\begin{align}
f &= - k_x \mathsf{e_x} - k_v \mathsf{e_v} + m \mathsf{g} + m \mathsf{\ddot{x}_d}, \\
M  &= k_R \mathsf{e_R} + k_\Omega \mathsf{e_\Omega} + \Omega \times \mathbf{J} \Omega
\enspace,
\end{align}
\end{subequations}
allowing one to achieve exponentially stable tracking behavior for initial attitude error less than $\sfrac{\pi}{2}$ (cf.~\cite{Lee}).

%The on-board controller that is used to track a small initial portion of the reference trajectory selected in each planning cycle (see Section~\ref{section:RHP} that follows) implements a customized version of control law \eqref{mav-control}. 
%The time interval associated with this selected small initial reference trajectory segment is referred to as the \emph{control horizon}.
%The length of the control horizon is dependent on the vehicle's speed, its computational capabilities, and its sensor update rate.

\section{Reactive Receding Horizon Planning}
\label{section:RHP}
%\subsection{Planner Objective}\label{problemstatement}

\subsection{Representing the Free Space}\label{graph}

%The key idea behind identifying a locally reachable portion of the free space is to present the \ac{fov} of the \acs{rgbd} sensor in the form of a finite, discrete set of points with fixed resolution, from which an \emph{open cover} of the point-cloud measurements has been removed along with all the occluded points.

Let $\mathcal{V} \in \mathbb{R}^3$ denote the visible space within the \ac{fov} of the \acs{rgbd} sensor. 
%$\mathcal{P} \subset \mathcal{V}$ a set of isolated points identified by the sensor's point-cloud, and $\mathcal{F} \subseteq \mathcal{V}$ be the representation of the obstacle-free space which the \ac{mav} can navigate in.
This \ac{fov} area contained in $\mathcal{V}$ is assumed to take the shape of a solid pyramid sector with its apex attached to the base frame of the sensor, with the depth direction of the sensor being aligned with the local $x$ (heading) frame axis of the \ac{mav}. 
The base of the \ac{fov} pyramid is outlined by the maximum range $R_\mathrm{max}$ of the sensor, while the side boundaries (polar and azimuth angles of the sector) are determined by the maximal viewing angles in the \ac{fov} of the sensor along the local $y$ and $z$ directions.
%rectangular pyramid that has its apex at the base frame attached to the sensor, its depth direction is aligned with the $x$ (heading) frame axis of the quadrotor, and the sides of the pyramid are determined by the fixed sensor's viewing angles in $y$ and $z$ directions. 
Denote $\phi_y$ and $\phi_z$ those maximum viewing angles of the sensor at its apex along the local $y$ and $z$ directions, respectively.
The motion planning process also takse as input a user-specified minimum range for perception, $R_\mathrm{min}$. 
Assuming now that the apex of the \ac{fov} pyramid is at local frame coordinates $(0, 0, 0)^\intercal$, any point within the \ac{fov} of the sensor can be expressed in the polar coordinates as $(r, \theta, \varphi)^\intercal$ where $R_\mathrm{min} \leq r \leq R_\mathrm{max}$, $-\phi_y \leq \theta \leq \phi_y$ and $-\phi_z \leq \varphi \leq \phi_z$.
By selecting a resolution $\delta r$ on the range and $\delta \theta$ on the viewing angles in both direction, the field of view of the sensor can be discretized and represented as an ensemble of points.
%This figure highlights three surfaces within the range of the sensor, on which the end points of the trajectories have been marked. 
%The average speed of the trajectory is different for different surfaces (please see section III-C for details). 
Each point in this ensemble represents a potential desired next location for the \ac{mav}, should a feasible path to this location exist. %The $(x, y, z)^\intercal$ coordinates of each point in the ensemble can be obtained in the sensor frame using polar to Cartesian coordinate transformation and can be further transformed in a global \emph{map} frame utilizing quadrotor's odometry and the transformation between the frame attached to quadrotor's \ac{cog} and the \acs{rgbd} sensor frame. 

%\vspace{-0.3cm}
\subsection{Reference Trajectory Generation}\label{pmp}

The \ac{cog} of the \ac{mav} is designated as the initial point of any candidate reference trajectory.
Using the (constant) transformation between \acs{cog} and sensor frame (at the \ac{fov} apex), the coordinates of all the points and rays can be expressed relative to body-fixed \acs{cog} frame  of the vehicle.

Given the ensemble of points within the field of view, a minimum snap trajectory to each of these point is constructed. 
Note that the dynamics of a quadrotor \ac{mav} enjoys differential flatness properties~\citep{Mellinger}, which ensure that all inputs and states can be written as functions of four (flat) outputs and their derivatives.
%associated with the vehicle's \acs{3d} position and yaw. 
The flat outputs are the Cartesian \ac{mav} position and its yaw angle, which can be brought together in a vector $(\mathsf{x},\psi)$.
Note that yaw $\psi$ is decoupled from $\mathsf{x}$ and can be steered independently.

The derivatives of $\mathsf{x}$ that are of interest are velocity $\mathsf{v}=\dot{\mathsf{x}}$, acceleration $\mathsf{a}=\ddot{\mathsf{x}}$, and jerk $\mathsf{j}=\dddot{\mathsf{x}}$.
In the flat output space, the \ac{mav} position dynamics can therefore be assumed to match those of a quadruple integrator with state vector $\mathsf{p}(t) = \left(\mathsf{x}^\intercal(t), \mathsf{v}^\intercal(t), \mathsf{a}^\intercal(t), \mathsf{j}^\intercal(t)\right)^\intercal$. 
The objective of the receding horizon planner now is to generate smooth trajectories $\mathsf{p}_{[N]}$ and $\mathsf{p}_{[N+1]}$, defined over the course of two consecutive planning cycles each of duration $\delta t$ and indexed $N$ and $N+1$,
\begin{align*}
\mathsf{p}_{[N]}(t, t\!+\!\delta t) 
&=\left(
\mathsf{x}_{d[N]}^\intercal, 
\,
\mathsf{v}_{d[N]}^\intercal,
\,
\mathsf{a}_{d[N]}^\intercal,
\,
\mathsf{j}_{d[N]}^\intercal
\right)^\intercal
\\
\mathsf{p}_{[N\!+\!1]}(t\!+\!\delta t, t\!+\!2\delta t) 
&\!=\!\left(
\mathsf{x}_{d[N\!+\!1]}^\intercal, 
\mathsf{v}_{d[N\!+\!1]}^\intercal,
\mathsf{a}_{d[N\!+\!1]}^\intercal,
\mathsf{j}_{d[N\!+\!1]}^\intercal
\right)^\intercal
\;,
\end{align*}
which always remain within $\mathcal{F}$ and  satisfy the boundary condition $\mathsf{p}_{[N]}(t+\delta t) = \mathsf{p}_{[N+1]}(t+\delta t)$, while being dynamically feasible, i.e.,  $f_\mathrm{min} \leq f \leq f_\mathrm{max}$ and $\|\Omega\| \leq \Omega_\mathrm{max}$.

Let $\mathsf{T}$ be the free trajectory terminal time (which will henseforth be referred to as the \emph{planning horizon}) and denote $\mathsf{p_0} = \left(\mathsf{x}_0^\intercal, \mathsf{v_0}^\intercal, \mathsf{a_0}^\intercal, \mathsf{j_0}^\intercal\right)^\intercal$,  $\mathsf{p_{\mathsf{T}}} = \left(\mathsf{x_T}^\intercal, \mathsf{v_T}^\intercal, \mathsf{a_T}^\intercal, \mathsf{j_T}^\intercal \right)^\intercal$ the trajectory boundary conditions at $t = 0$ and $t = \mathsf{T}$, respectively. 
Then let $\mathsf{u}(t) = \od[4]{\mathsf{x}(t)}{t} $ denote  snap, and treat it as the control input for the flat space dynamics. 
For a \emph{relative weight parameter} $k\in \mathbb{R}_{+}$, the free terminal time optimal control problem the solution of which are the desired reference trajectories, is formulated as:
\begin{equation}
 \hspace{-0.2em}   \begin{cases}
    \min \int_{0}^{\mathsf{T}} \big( k + \frac{1}{2}\norm{\mathsf{u}(t)}^2 \big) \dif t \qquad 
    \text{subject to}
  \\
        \dot{\mathsf{x}}(t) = \mathsf{v}(t), \enspace 
        \dot{\mathsf{v}}(t) = \mathsf{a}(t), \enspace
        \dot{\mathsf{a}}(t) = \mathsf{j}(t), \enspace
        \dot{\mathsf{j}}(t) = \mathsf{u}(t)\\
       \mathsf{p}(0) = \mathsf{p_0}, \enspace \mathsf{p}(\mathsf{T}) = \mathsf{p_{T}} 
       \enspace.
    \end{cases}  
    \label{minsnap}
\end{equation}
The cost function of the optimal control problem \eqref{minsnap} is a linear combination of two performance factors: 
\begin{inparaenum}[(i)] 
\item the incremental cost associated with the terminal time (time optimality), captured by the constant integrand term; and 
\item the incremental cost that penalizes the cummulative magnitude of  snap $\mathsf{u}$ along the trajectory. 
\end{inparaenum}
By tuning $k$, one can adjust how  aggressive the reference trajectory obtained is. 
Smaller values for $k$ place a smaller penalty on tracking time and therefore result to slower motion. 
\acs{gazebo} simulation data at various $k$ and speeds had been collected to fit a relation between the maximum speed along a candidate trajectory and parameter $k$. 
This relation comes out to be an interpolated curve of the form $\mathsf{v}_\mathrm{candidate} = \alpha\, k^{1/\beta}$ and has been found to work well in practical scenarios to guide the selection of the cost weight $k$ based on the maximum robot speed afforded for the mission at hand.
The cost weight $k$ is particular to the candidate trajectory and varies for the different candidate trajectories in the ensemble, since the maximum speed afforded along a candidate trajectory $\mathsf{v}_\mathrm{candidate}$ itself varies in proportion to the ratio of the candidate trajectory's length to the length of the longest candidate trajectory. 
This leads to smaller trajectories having lower top speeds, making the \ac{mav} automatically slow down in the vicinity of dense obstacles. 

Denoting $t$ the time elapsed since the start of the whole planned maneuver, $d$ the vehicle's remaining distance to its goal, $r$ being the distance of the point in the ensemble from the camera focal point,  $v_\mathrm{max}$ the desired maximum \ac{mav} speed, 
and $k_t$ and $k_d$ being positive tuning parameters,
the velocity used to calculate the weighing factor $k$ in \eqref{minsnap} to generate a particular candidate trajectory is given by\useshortskip
\begin{equation} \label{sigmoid}
\mathsf{v_{candidate}} = \erf{(k_t \, t)} \erf{( k_d \, d )} \left(\tfrac{r}{R_{max}}\right)\, v_\mathrm{max}
\enspace.
\end{equation}
Compared to alternative trapezoid velocity profiles~\citep{Mellinger}, \eqref{sigmoid} produces a velocity profile for the entire remaining quadrotor trajectory that is also effective in scenarios involving moving target interception, in which the vehicle needs to adjust its speed to match that of its target while at the vicinity of the latter. 
A more detailed discussion of the effect of this velocity profile on tracking dynamic targets follows in the section on target tracking.

%Figure~\ref{Fig4} illustrates this dependence showing the maximum speed achieved along the trajectory as a function of $k$, as an interpolated curve of the form $\mathsf{v}_\mathrm{max} = \alpha\, k^{1/\beta}$ (Fig.~\ref{Fig4}).
%Such a relationship can guide the selection of the cost weight $k$ based on the maximum robot speed afforded for the mission at hand.

%\begin{figure}[h!]
  %  \vspace{-0.3cm}
%	\centering
%	\includegraphics[width=0.75\linewidth]{velocity_vs_k.png}
%	\caption{
%	Relation between the cost weight $k$ and maximum speed along the vehicle's reference trajectory. Points mark simulation data; the dashed curve is a polynomial fit with $\alpha = 11,354.1 $ and $\beta = 10.1143$.
%	\label{Fig4}
%	\vspace{-0.2cm}
%	}
%\end{figure}

To solve \eqref{minsnap} one can utilize Pontryagin's Minimum principle~\citep[Chapter 6]{michaelathens} as follows.
Let $\mathsf{\lambda_x}$, $\mathsf{\lambda_v}$, $\mathsf{\lambda_a}$ and $\mathsf{\lambda_j}$ be the \emph{costate} vectors. 
Each such vector has three components, one for each spatial direction $x$, $y$, and $z$. 
%such that each $\mathsf{\lambda_{\cdot}} = [\mathsf{\lambda_{\cdot x}}, \mathsf{\lambda_{\cdot y}}, \mathsf{\lambda_{\cdot z}}]$ is a 3 dimensional vector in $x$ $y$ and $z$ direction, 
Let $\langle \cdot,\cdot \rangle$ denote the standard inner product between of two vectors, and express the Hamiltonian $\mathsf{H}$ for this problem as 
\vspace{-0.15cm}
\begin{equation} \label{hamiltonian}
\mathsf{H} = k + \frac{1}{2}\norm{\mathsf{u}}^2 +  
\langle \mathsf{\lambda_x}, \mathsf{v}\rangle + 
\langle \mathsf{\lambda_v}, \mathsf{a}\rangle +  
\langle \mathsf{\lambda_a}, \mathsf{j} \rangle +
\langle \mathsf{\lambda_j}, \mathsf{u} \rangle
\enspace.
\end{equation}
The optimal solution is derived from the condition $\mathsf{H}(\mathsf{x^{*}}, \mathsf{u^{*}}, \mathsf{\lambda^{*}}, t) \leq \mathsf{H}(\mathsf{x^{*}}, \mathsf{u}, \mathsf{\lambda^{*}}, t)$, which, since the Hamitonian is quadratic in the control inputs, leads to
\begin{equation}
\pd{\mathsf{H}}{\mathsf{u}} = 0 \implies 
\begin{cases}
\mathsf{u}_x = -\lambda_{\mathsf{j}x}
        & \\
\mathsf{u}_y = -\lambda_{\mathsf{j}y} & \\
\mathsf{u}_z = -\lambda_{\mathsf{j}z}
\enspace.
\end{cases}  
\label{hamiltonian_minimization}
\end{equation}
The costate dynamics now is
%$$\enspace \dot{\mathsf{\lambda}_p} = - \frac{\partial \mathsf{H}}{\partial {\mathsf{p}}} = 0, \enspace \enspace \dot{\mathsf{\lambda}_v} = - \frac{\partial \mathsf{H}}{\partial {\mathsf{v}}} = -\mathsf{\lambda_p}, \enspace \dot{\mathsf{\lambda}_a} = - \frac{\partial \mathsf{H}}{\partial {\mathsf{a}}} = -\mathsf{\lambda_v}$$

%$$\mathsf{
%\dot{\lambda}_p = - \frac{\partial H}{\partial {p}} = 0, \enspace 
%\dot{\lambda}_v = - \frac{\partial H}{\partial {v}} = -\lambda_p, \enspace
%\dot{\lambda}_a = - \frac{\partial H}{\partial {a}} = -\lambda_v, \enspace
%\dot{\lambda}_j = - \frac{\partial H}{\partial {j}} = -\lambda_a
%} $$

%\begin{multline*}
%\mathsf{
%\dot{\lambda}_p = - \frac{\partial H}{\partial {p}} = 0, \enspace 
%\dot{\lambda}_v = - \frac{\partial H}{\partial {v}} = -\lambda_p}\\
%\mathsf{
%\dot{\lambda}_a = - \frac{\partial H}{\partial {a}} = -\lambda_v, \enspace
%\dot{\lambda}_j = - \frac{\partial H}{\partial {j}} = -\lambda_a
%}
%\end{multline*}

\begin{align*}
 \dot{\lambda}_\mathsf{x} 
 &= 
 - \pd{\mathsf{H}}{\mathsf{x}} = \mathsf{0} 
 &
 \dot{\lambda}_\mathsf{v} 
 &= 
 - \pd{\mathsf{H}}{\mathsf{v}} = -\lambda_\mathsf{p}
 \\   
 \dot{\lambda}_\mathsf{a} 
 &= 
 - \pd{ \mathsf{H}}{\mathsf{a}} = -\lambda_\mathsf{v}
 &
 \dot{\lambda}_\mathsf{j} 
 &= 
 - \pd{ \mathsf{H}}{\mathsf{j}} = -\lambda_\mathsf{a} 
 \enspace.
\end{align*}  
%    \label{min-snap-costate}

Fortunately, this problem affords independent solution along each direction.
The following equation set, where $c = (c_{x0}, \cdots , c_{x7})$ denotes an $8 \times 1$ vector of constant coefficients, illustrates the solution along in $x$ direction; the other two directions feature identical polynomials: 
\begin{equation}
\hspace{-0.7em}
\left.
\begin{array}{l} 
\lambda_{\mathsf{p}x} 
    = c_{x7} \\
\lambda_{\mathsf{v}x} 
    = -c_{x7} t + c_{x6}\\
\lambda_{\mathsf{a}x} 
    = c_{x7} \frac{t^2}{2} - c_{x6} t + c_{x5} \\
\mathsf{u}_x 
    =  c_{x7}\frac{t^3}{6} - c_{x6} \frac{t^2}{2} + c_{x5} t - c_{x4}\\
\mathsf{j}_x 
    = c_{x7}\frac{t^4}{24} - c_{x6} \frac{t^3}{6} + c_{x5} \frac{t^2}{2} - c_{x4} t + c_{x3}\\
\mathsf{a}_x 
    = c_{x7}\frac{t^5}{120} - c_{x6} \frac{t^4}{24} + c_{x5} \frac{t^3}{6} - c_{x4}  \frac{t^2}{2} + c_{x3} t + c_{x2}\\
\mathsf{v}_x 
    = c_{x7}\frac{t^6}{720} - c_{x6} \frac{t^5}{120} + c_{x5} \frac{t^4}{24} - c_{x4}  \frac{t^3}{6} + c_{x3} \frac{t^2}{2} \\
    \qquad + c_{x2} t + c_{x1}\\
\mathsf{x}_x 
    = c_{x7}\frac{t^7}{5040} - c_{x6} \frac{t^6}{720} + c_{x5} \frac{t^5}{120} - c_{x4}  \frac{t^4}{24} + c_{x3} \frac{t^3}{6}  \\
    \qquad + c_{x2} \frac{t^2}{2} + c_{x1} t + c_{x0} \enspace.
\end{array}  \right\}  
    \label{min-snap-sol}
\end{equation}
The optimal trajectory, therefore, is a 7\textsuperscript{th} order polynomial in time. 
Enforcing the boundary conditions at $t = 0$ gives $c_{x0} = \mathsf{p}_{x0}$, $c_{x1} = \mathsf{v}_{x0}$, $c_{x2} = \mathsf{a}_{x0}$ and $c_{x3} = \mathsf{j}_{x0}$, while 
the remaining coefficients, $c_{x4}$ through $c_{x7}$, are obtained from the boundary conditions at $t = \mathsf{T}$:
\begin{equation*}
\label{three_coeff}
\left[
\begin{smallmatrix}
c_{x7}\\ 
c_{x6}\\ 
c_{x5}\\ 
c_{x4}
\end{smallmatrix}
\right]
= 
\medmath{
        %\begin{bmatrix}
        %   c_{x7} \\
        %  c_{x6} \\
        %   c_{x5}\\
        %   c_{x4}
        %\end{bmatrix} =
        \left[
        \begin{smallmatrix}
           \frac{\sf T^7}{5040} \enspace \frac{\sf -T^6}{720} \enspace \frac{\sf T^5}{120} \enspace \frac{\sf -T^4}{24} \\
           \frac{\sf T^6}{720} \enspace \frac{\sf -T^5}{120} \enspace \frac{\sf T^4}{24} \enspace \frac{\sf -T^3}{6}\\
           \frac{\sf T^5}{120} \enspace \frac{\sf -T^4}{24} \enspace \frac{\sf T^3}{6} \enspace \frac{\sf -T^2}{2}\\
           \frac{\sf T^4}{24} \enspace \frac{\sf -T^3}{6} \enspace \frac{\sf T^2}{2} \enspace \sf -T\\
        \end{smallmatrix}
        \right]^{-1}  
        \hspace{-1.5em}
        \left[   
        \begin{smallmatrix}
     \mathsf{p}_{x\mathsf{T}} - (\mathsf{p}_{x0} + \mathsf{v}_{x0} \mathsf{T}+ \frac{1}{2} \mathsf{a}_{x0} \mathsf{T}^2) + \frac{1}{6} \mathsf{j}_{x0} \mathsf{T}^3\\
 \mathsf{v}_{x\mathsf{T}} - (\mathsf{v}_{x0} + \mathsf{a}_{x0} \mathsf{T} + \frac{1}{2} \mathsf{j}_{x0} \mathsf{T}^2) \\
      \mathsf{a}_{x\mathsf{T}} - (\mathsf{a}_{x0} + \mathsf{j}_{x0} \mathsf{T})\\
           \mathsf{j}_{x\mathsf{T}}- \mathsf{j}_{x0}
        \end{smallmatrix}
        \right]
        }
\end{equation*}
expressing the optimal trajectory coefficients as a function of the (yet unknown) free terminal time $\mathsf{T}$. 

The free terminal time $\mathsf{T}$ can be determined as follows.
With the $c_x$ trajectory coefficients are explicitly expressed in terms of $\mathsf{T}$, one substitutes and writes the control input at the terminal time as
\[
\mathsf{u}_{xT} 
= 
\frac 
 {840 (\mathsf{p}_{x0}
 -\mathsf{p}_{x\mathsf{T}})}
 {\mathsf{T}^4} +
\frac {360 \mathsf{v}_{x0}} 
 {\mathsf{T}^3} +
\frac {60 \mathsf{a}_{x0}} 
 {\mathsf{T}^2} + 
\frac {4\mathsf{j}_{x0}}
 {\mathsf{T}} 
\]
(control inputs at $\mathsf{T}$ in $y$ and $z$ are similarly obtained).
Velocity, acceleration and jerk at time $\mathsf{T}$ are all set to zero, while  
the transversality condition~\citep{michaelathens} at $\mathsf{T}$ requires $\mathsf{H(T)} = 0$.
Combining these conditions, with \eqref{hamiltonian_minimization} results to 
\[
    k + \frac{1}{2} \norm{\mathsf{u_T}}^2 + \langle \mathsf{\lambda_T}, \mathsf{u_T}\rangle = 0 \Longrightarrow \norm{\mathsf{u_T}}^2 = 2k
    \enspace.
\]
This is essentially an 8\textsuperscript{th} degree polynomial equation which can now produce $\mathsf{T}$.
Indeed, if for the sake of brevity we set
\begin{align*}
l_1 \!\triangleq 
840 (p_{x0}\!-\!p_{x\mathsf{T}})
& &
m_1 \!\triangleq 
360\, \mathsf{v}_{x0}
& &
n_1 \!\triangleq \!
60 \,\mathsf{a}_{x0}
& &
o_1 \!\triangleq \!
4\,\mathsf{j}_{x0}
\end{align*}
(work similarly for $y$ and $z$ using indices $2$ and $3$, respectively), and then substitute back we obtain the polynomial equation
\begin{multline}
 -2k \mathsf{T}^8 + 
  \sum_{i = 0}^{3} o^2_i \, \mathsf{T}^6 + 
 2 \sum_{i = 0}^{3} n_i o_i
 \, \mathsf{T}^5 \\
+ \sum_{i=0}^{3} (n^2_i + 2m_i o_i) \, \mathsf{T}^4 + 
2 \sum_{i = 0}^{3} (l_i o_i+ m_i n_i) \, \mathsf{T}^3  \\
+ \sum_{i=0}^{3} (m^2_i + 2l_i n_i) \, \mathsf{T}^2 
+ 2 \sum_{i = 0}^{3} l_i m_i \, \mathsf{T} + \sum_{i = 0}^{3} l^2_i = 0 
\enspace.
\label{time_poly}
\end{multline}
This equation
can be efficiently solved numerically, returning three pairs of complex conjugate roots and a single pair of real roots, of which one is negative and the other is positive---the only acceptable root.

The vehicle's yaw angle $\psi$ is dynamically decoupled from its position.
A reference trajectory for yaw can be constructed in a similar way, given that the relative degree of flat output $\psi$ is two.
In the flat output space, therefore, the yaw dynamics can be expressed as as a double integrator.
Let us define the yaw state vector $\Psi \triangleq (\psi,\dot{\psi})^\intercal$, treating $w\triangleq\ddot{\psi}$ as the virtual control input for this dynamical subsystem.
For the yaw trajectory we have boundary conditions on both sides, denoted $\Psi(0) = \Psi_0 \triangleq (\psi_0, \dot{\psi}_0)$ and $\Psi(\mathsf{T}) = \Psi_\mathsf{T} \triangleq (\psi_\mathsf{T}, \dot{\psi}_\mathsf{T})$.
The reference yaw trajectory is obtained as a solution of
\begin{equation}
    \begin{cases}
    \min \frac{1}{2}
    \int_{0}^{\mathsf{T}} w(s)^2 \dif s\\
 \text{subject to} \\
    \Psi = (\psi,\dot{\psi})\,,
    \enspace
    \ddot{\psi}(t) = w(t)\\
       \Psi(0) = \Psi_0, \enspace \Psi(\mathsf{T}) = \Psi_{\mathsf{T}} 
       \enspace,
    \end{cases}  
    \label{min-yaw}
\end{equation}
which is a 3\textsuperscript{rd} order polynomial of the form 
\[
\psi(t) = 
\gamma_0 + \gamma_1 \, t 
+ \gamma_2 \, t^2 
+ \gamma_3 \, t^3 
\enspace,
\]
with coefficients given as
\begin{align*}
\gamma_0 &= \psi_0 & 
\gamma_2 &= \textstyle \frac{1}{2}\left(\frac
    {6(\psi_\mathsf{T} - \psi_0)} {\mathsf{T}^2} - \frac{2
    (\dot{\psi}_{\mathsf{T}} + 2\dot{\psi}_{0})
    }{\mathsf{T}} \right)
\\    
\gamma_1 &= \dot{\psi}_{0} &
\gamma_3 &= \textstyle
\frac{1}{6}\left(
     \frac{6(\dot{\psi}_\mathsf{T} + \dot{\psi}_0)}{\mathsf{T}^2} -\frac{12(\psi_\mathsf{T} - \psi_0)}{\mathsf{T}^3} \right)
     \enspace.
\label{min-snap}     
\end{align*}

%The trajectory in the other two flat outputs can also be obtained similarly using the same terminal time $\mathsf{T}$ . The total energy can then be obtained since the control input vector $\mathsf{u} (t)$ along the trajectory is known.

\begin{figure}[h!]
	\centering
	\includegraphics[width=0.8\linewidth]{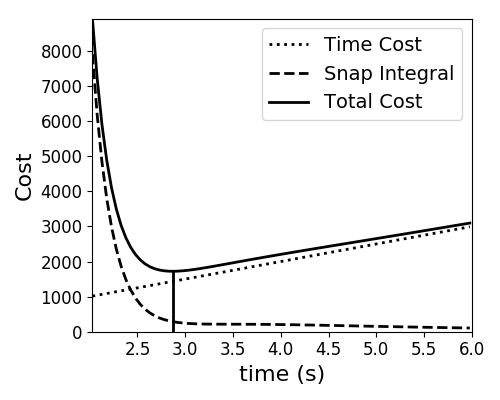}
	\caption{Temporal evolution of the optimization cost functional  over time, showing how the variation of its two affine cost components contribute to its final value and yield a convex curve with respect to time. 
		\label{Fig3}
		%\vspace{-0.5cm}
		}	
\end{figure}

Figure~\ref{Fig3} shows a sample evolution of the cost functional in \eqref{minsnap} for some particular set of boundary conditions. 
%\todo[inline]{\small any chance we can produce this graph with different linetypes instead of colors?
%then type "Cost" vertically and "time (s)" horizontally. Done}
The cost associated with the terminal time (dotted curve) increases while the integral of the snap (dashed curve) reduces with time. The vertical line represents the positive real root $\mathsf{T}$ of \eqref{time_poly} which minimizes the total cost (solid curve). 

%By tuning $k$ one regulates the aggressiveness of the reference trajectory.

%The trajectory generated using the optimal $\mathsf{T}$ in \eqref{min-jerk-sol} could be of varying aggressiveness depending upon the value of parameter $k$ in \eqref{minsnap}. A smaller value of $k$ would return a trajectory which is safe but too slow and a higher value of $k$ could return the trajectory that is not dynamically feasible. 

In the context of the particular case study considered here, the best radiation detection performance has been shown to be achieved when the robot closes the distance between its sensor and the source as fast as possible~\citep{AURO-Jianxin,Yadav}.
This implies that aggressive reference trajectories are preferable, motivating us to select $k$ based on the maximum speed limitations of the \ac{mav}. 
With this in mind, the methodology outlined above can thus provide motion plans that would be 
\begin{inparaenum}[(a)]
\item feasible given the dynamic constraints of the vehicle, 
\item as aggressive as possible,  and 
\item high-performance in terms of  radiation detection accuracy. 
\end{inparaenum}

\begin{figure}[h!]
	\centering
	\includegraphics[width=1.0\linewidth]{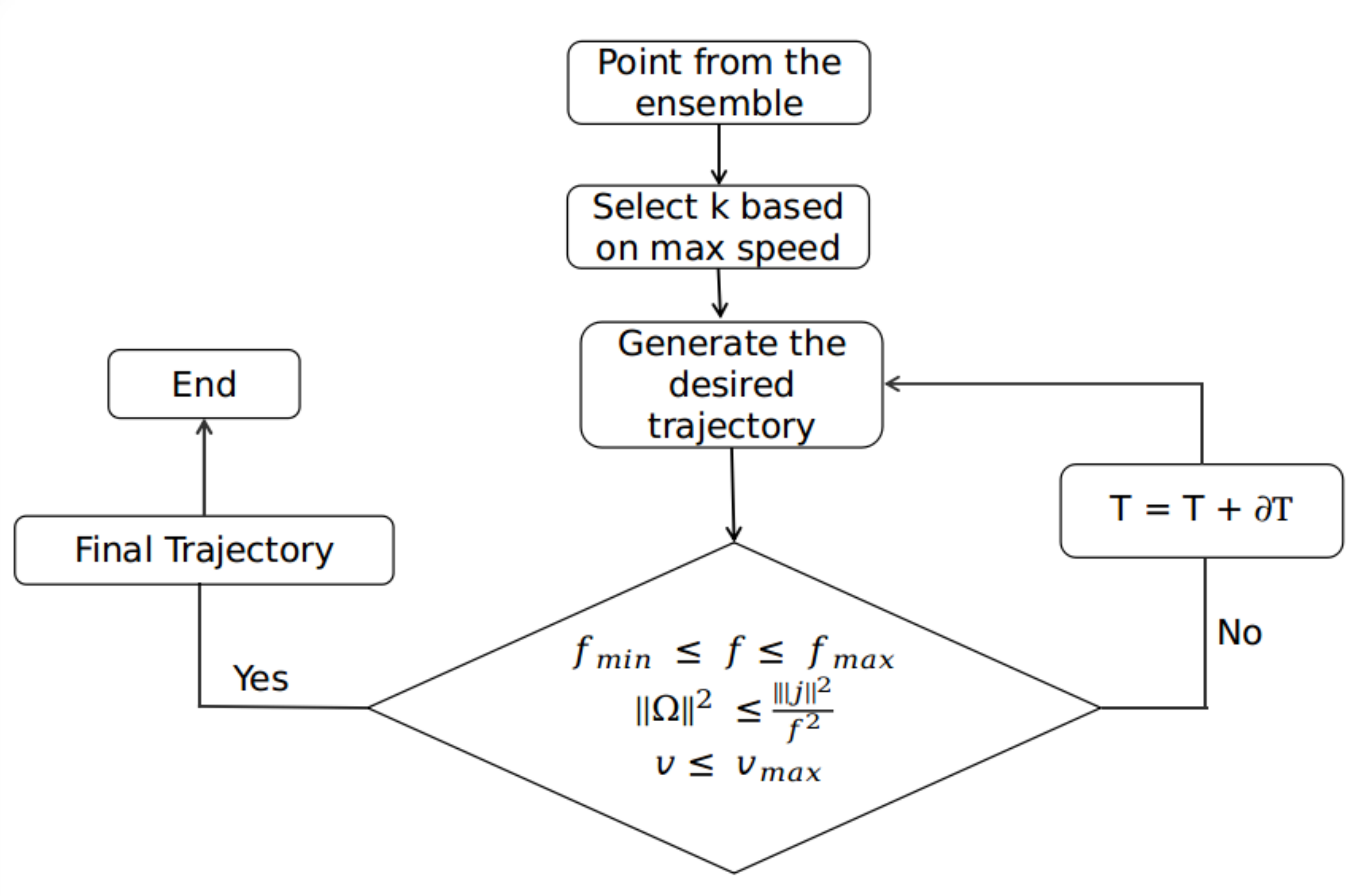}
	\caption{
	The decision block diagram for the reference trajectory generation process. 
	If the actuation limit check fails, the terminal time $\mathsf{T}$ is increased slightly by some $\delta \mathsf{T}$ and the trajectory is regenerated.
	\label{Fig5}
	%\vspace{-1.0cm}
	}
\end{figure}

While the reference trajectory is dynamically feasible by design, its conformity to actuation constraints is verified after its generation~\citep{MWMueller} (see Fig.~\ref{Fig5}).
%We therefore check~\cite{MWMueller} if the trajectory is dynamically feasible and depending upon this check allow a small perturbation in the terminal time $\mathsf{T}$ to get a more aggressive trajectory or a safer one. 
Here, we have actuation (upper)  bounds on the magnitude of the input thrust $f = m \norm{\mathsf{a} - \mathsf{g}}$ and on the square of the sum of roll and pitch angular velocities in terms of jerk and thrust,  $\sfrac{\norm{\mathsf{j}}^2}{f^2}$. 
%The latter translates to a constraint on admissible angular velocities. 
These actuation constraints stem from
\begin{inparaenum}[(a)]
\item equipment limitations, and 
\item the maximum deceleration that the vehicle can undergo during an emergency stopping maneuver. 
\end{inparaenum}
In the reactive navigation architecture described here, emergency stopping maneuvers are engaged when the \ac{mav} cannot find a safe path in its free workspace. 
Finally, a linear velocity constraint is imposed in order to reduce motion blur which would otherwise affect the robot's \ac{vio}.

%\todo[inline]{Is it possible to have the figure in pdf form? Done.}

\subsection{Local Goal and Collision Costs}\label{cost}

Once the ensemble of candidate reference trajectories is generated, the candidate trajectories that intersect with any obstacle is excluded from the ensemble. 
The methodology is motivated from earlier work of \cite{yadav_iros}, and has been modified to suit the new trajectory generation methodology.

The obstacle point cloud $\mathcal{P}$ is first converted into a \acs{kd}-Tree and each trajectory is discretized and represented in the form of a finite (size $n$) sequence of points. Thereafter a query is run to find points of the \acs{kd}-tree that lies within a ball of radius $r$ from each of these points on the candidate trajectory. 
Parameter $r$ is chosen so that it can fully enclose the quadrotor, with a suitable---based on how conservative with respect to the practical risk collision due to uncertainty or disturbances---safety margin. 
A candidate trajectory is said to be intersecting with an obstacle if any point (among the n points in which it is discretized) on it has any point of \acs{kd}-tree within this ball of radius $r$; such a trajectory is excluded from the ensemble.

%Once the ensemble of candidate reference trajectories is generated, we select balls of radius $r$ around each point $p\in \mathcal{P}$ in the sensor's point cloud, 
%Parameter $r$ is chosen so that it can fully enclose the quadrotor, with a suitable---based on how conservative with respect to the practical risk collision due to uncertainty or disturbances---safety margin. 
%The process described next is employed to exclude from further consideration candidate trajectories that intersect with any such ball. The methodology described is motivated from our previous work~\cite{yadav_iros} and has been modified to suit the new trajectory generation methodology. 

%The obstacle point cloud $\mathcal{P}$ is first converted into a \acs{kd}-Tree and each trajectory is discretized and represented in the form of a finite (size $n$) sequence of points.
%The sequences themselves are represented in another \acs{kd}-Tree, that now represents all candidate  trajectory discretizations, and based on those two \acs{kd}-Trees, a query is run for nearest neighbor pairs; those sequences with points within $r$ of point cloud points are associated to candidate trajectories which are discarded.
%The ray \acs{kd}-Tree can now be queried from the pointcloud \acs{kd}-Tree to determine nearest neighbours, and the rays intersecting with $r$-balls around pointcloud points---those rays are subsequently removed. 

\begin{figure*}[t]
    \centering
    \subfloat[Distance to Goal Cost]{%
    \centering
    \includegraphics[width=0.33\linewidth]{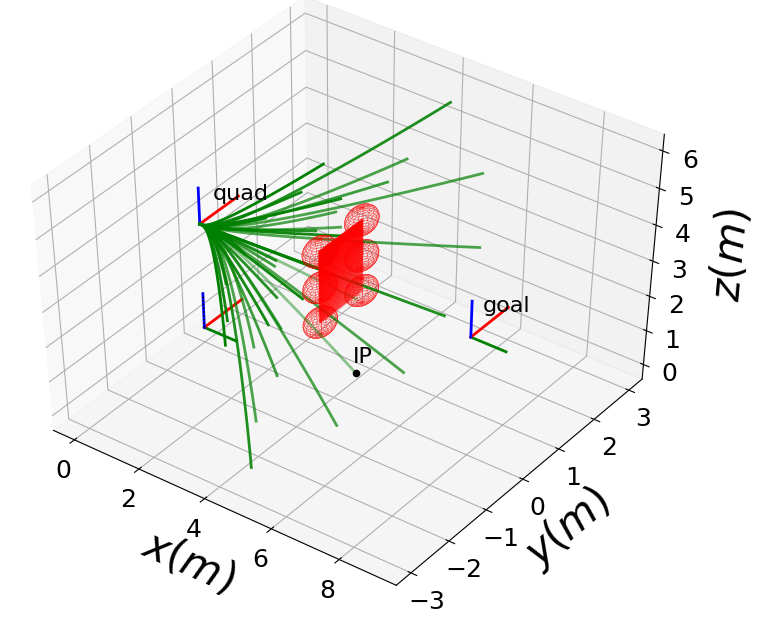}
    \label{Fig7a}
    }
    \subfloat[Collision Cost]{%
    \centering
    \includegraphics[width=0.33\linewidth]{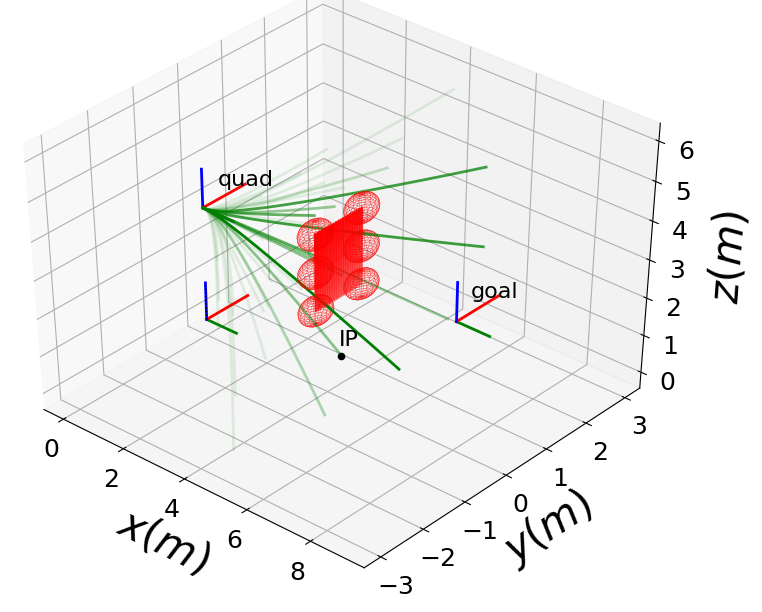}
    \label{Fig7b}
    }
    \subfloat[Total Cost]{%
    \centering
    \includegraphics[width=0.33\linewidth]{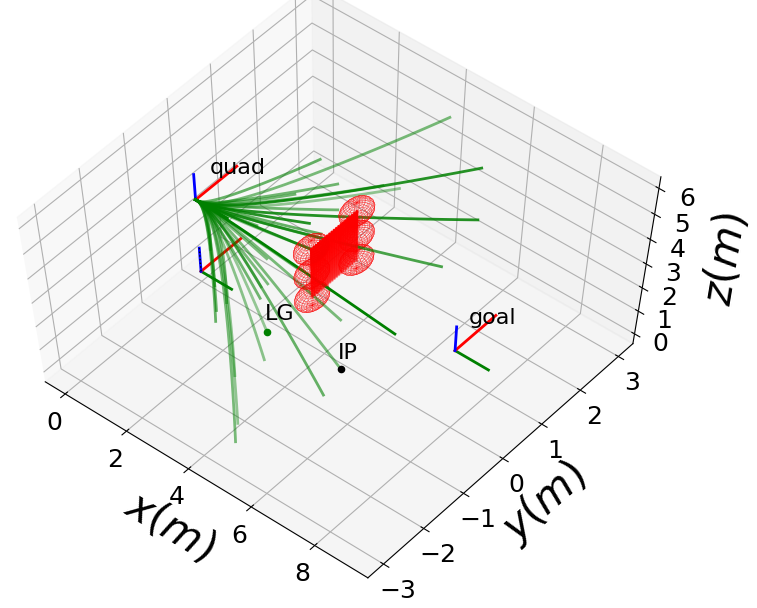}
    \label{Fig7c}
    }
    \caption{Assigning costs to collision-free trajectories. The trajectories with high costs are plotted as darker green lines. The trajectory with minimum total cost is selected as the final trajectory through which the robot traverses. The corresponding end point is the \emph{local goal}.
    \label{Fig7}
    %\vspace{-0.5cm}
    }
\end{figure*}

Among the collision-free candidate trajectories, the optimal one should strike an acceptable balance between safety (against collision), and speed of convergence to the goal point.
An illustration of the subsequent process of selecting this optimal reference trajectory is given in Fig.~\ref{Fig7} (for details on the associated computational requirements, see  Section~\ref{results}). 
In lieu of a global planner which would otherwise sketch a complete path from initial to final configurations, the reactive local planner defines an \emph{intermediate point} as the point in the (collision free) ensemble closest to the final goal (denoted \textsf{IP} in Fig.~\ref{Fig7a}. 
It then assigns a cost to each trajectory, in the form of a linear weighted sum of two cost components: the first component is based on the distance of each trajectory end-point to the intermediate point, normalized over the maximum distance; the second component is a normalized collision cost that captures how close the trajectory comes to $\mathcal{P}$.

%\todo[inline]{\small The following passage talks about rays instead of trajectories; I'm guessing this is a leftover from earlier draft, yes, it is corrected.}

Denote the total number of collision-free trajectories and the Euclidean distance between the end point of the $i$\textsuperscript{th} trajectory and the intermediate point, $p$ and $d_i$, respectively. 
Set $d_\mathrm{max} \triangleq \max_{i} d_i$, and let $\hat{r} \ge r$ be an additional safety margin (over the radius around detected obstacles). 
The minimum distance $\rho_i$ of trajectory $i$ to obstacles is found by quering a \acs{kd}-Tree over $\mathcal{P}$ and minimizing over the query response. 
With these data at hand, the collision cost for trajectory $i$ is found as\useshortskip
\[
c_{\mathrm{coll}_i} =
    \begin{cases}
    \frac{1+\hat{r}^4}{\hat{r}^4} \cdot \frac{[(\rho_i-r)^2-\hat{r}^2]^2}{ 1 + [(\rho_i-r)^2-\hat{r}^2]^2} & \text{if} \enspace \rho_i - r \le \hat{r}\\
    0 & \text{otherwise\;.} 
    \end{cases}
\]

%\begin{fralign*}
%&c_i
%= k_1 \tfrac{d_i}{d_{max}}+ %k_2 \,c_{\mathrm{coll}_i}, \\
%&c_{coll_i} =
%    \begin{cases}
%    f \cdot \frac{((\delta_i-r)^2-\hat{r}^2)^2}{ 1 + ((\delta_i-r)^2-\hat{r}^2)^2}, \enspace \text{if} \enspace \delta_i - r \le \hat{r}\\
%    0 \enspace \enspace \text{otherwise}
%    \end{cases}, \enspace f = \frac{1+\hat{r}^4}{\hat{r}^4}\\
%&f = \frac{1+\hat{r}^4}{\hat{r}^4},
%\end{fralign*}

The collision cost function normalizes the cost of each trajectory into the $[0,1]$ interval.
Thus any trajectory that touches a ball around its nearest obstacle is assigned a collision cost of 1, while any trajectory that lies at least $\hat{r}$-away from every obstacle incurs zero cost.
All other trajectories are assigned costs within the $(0, 1)$ interval. 
The end point of the trajectory with the lowest total cost becomes the  \emph{local goal} (i.e., within the \ac{fov}) for the planner (marked with a green dot in Fig.~\ref{Fig7c}).

Selecting positive weights $k_1,\,k_2 \in (0,1)$, the cost contributions of trajectory $i \in \{0,\ldots, p\}$ due to its proximity to the intermediate point and obstacles are combined into an aggregate cost expression
\[
c_i
= k_1 \tfrac{d_i}{d_\mathrm{max}}+ k_2 \,c_{\mathrm{coll}_i}
\enspace.
\]
The trajectory with the least such cost is selected as the reference trajectory between the robot's current position and the local goal.

%\todo[inline]{\small Can you remove the double parentheses from the subfigure captions?}

%\subsection{Receding Horizon Trajectory Tracking}\label{Planning}

%\begin{figure}[h!]
%	\centering
%	\includegraphics[width=0.75\linewidth]{goal_chase6.png}
%	\caption{Receding horizon trajectory generation. Dashed green lines shows generated trajectories wile solid green is concatenated trajectory.    
%		\label{Fig8}
%	}
%\end{figure} 

%specifically, the control horizon should be longer than the \emph{sensor update horizon} (the period between sensor updates) plus some safety margin. 
%the computation time needed to solve \eqref{min-snap}.
The \acs{mav} tracks a small portion of the reference trajectory, hereafter referred to as the \emph{control horizon}. The length of the control horizon is dependent on the vehicle's speed, its computational capabilities, and its sensor update rate.
Each time the \acs{mav} receives a sensor update, 
%while executing a pre-computed reference trajectory, 
it generates a new reference trajectory (segment) and appends it to the end of the segment it is currently tracking.
By design, the transition between subsequent reference trajectory segments is smooth.
%and starts executing the newly computed one in anticipation of the next sensor update.
The process in between sensor updates constitutes a \emph{replanning cycle}.
This process is illustrated in Fig.~\ref{Fig-rhp} which shows different trajectories generated from the starting point until the end point. 
%for the scenario of Fig.~\ref{Fig7}.
The complete implementation for planning, control and state estimation is open-source.\footnote{ {\color{Blue}https://github.com/indsy123/quad\_navigation\_and\_target\_tracking}}

% this picture below is actually better for figure 8
%\begin{figure*}
%  \includegraphics[width=\textwidth,height=4cm]{trajectory_rviz.png}
%  \caption{This is a test.}
%\end{figure*}

\section{Target Tracking}\label{targettracking}
\subsection{Object Detection based on \acs{ssd}-MobileNetV2}\label{dnn}
%{\color{red} Can write about the architecture of \acs{ssd}-MobilenNetV2 with a picture?}
 
The motivating application for this work is the detection of weak radiological material on moving ground vehicles, and for this to be achieved via aerial means of sensing and measurement in a timely manner, the \acp{mav} need to get close to their target as quickly as possible.
To this end, the methodology of Section~\ref{section:RHP} is extended to dynamic target interception and tracking scenarios.
Necessary extensions include the capability of the \ac{mav} to autonomously detect its target within its \ac{fov}, and estimate its target's relative position and speed based on visual means.
%The word detection in Section-II refers to the this decision making while in this section it sometimes refers to a classification  
For this task an open source implementation of MobileNetV2~\citep{Sandler} deep neural network in TensorFlow~\citep{abadi2016tensorflow} is utilized and combined with \ac{ssd}~\citep{Liu_2016} for object detection.
%The MobileNetV2  module  takes as an input a low-dimensional compressed representation which is first expanded to high dimension and filtered with a lightweight depthwise convolution. Features are subsequently projected back to a low-dimensional representation with a linear convolution. Features extracted by MobileNetV2 are utilized in a \ac{ssd} architecture that uses a single feed-forward convolutional network to directly predict classes and anchor offsets without requiring a second stage per-proposal classification operation, thereby achieving higher computation speed even in resource constrained hardware. 
%Transfer learning can also be used to adapt an existing pre-trained neural network to perform the desired object detection.

%Open source implementation of \acs{ssd}-MobileNetV2 in TensorFlow~\cite{abadi2016tensorflow} has been utilized for object detection. 

\begin{figure*}[t]
\centering
\subfloat[Detection+Feature Matching]{%
  \includegraphics[width=0.4\textwidth]{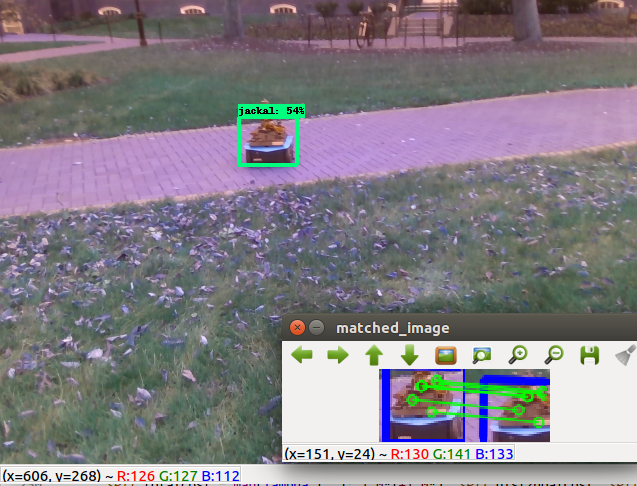}%
  \label{target-tracking-a}
}
\subfloat[Workflow]{%
    \centering
  \includegraphics[width=0.4\textwidth]{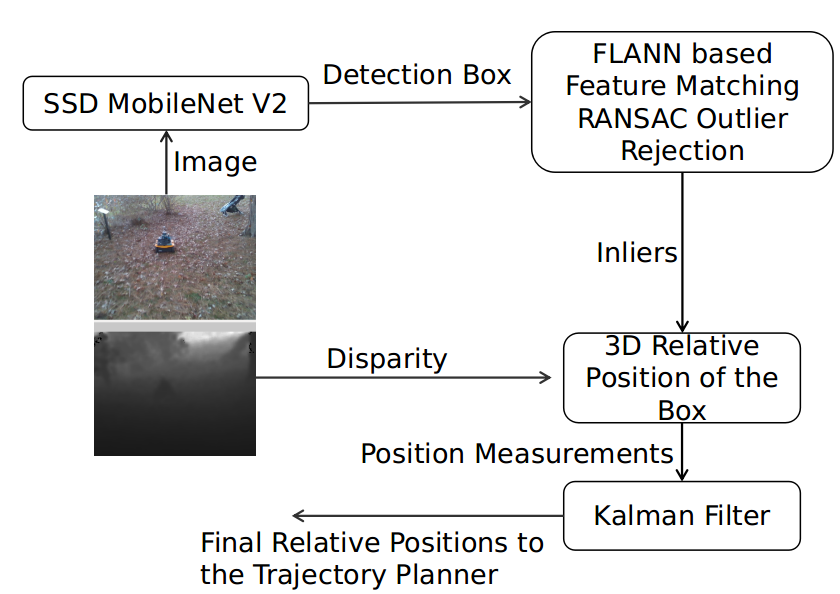}%
  \label{target-tracking-b}
}
\caption{
\subref{target-tracking-a}) The mobile target in the experimental study. The detection box is shown in green while the image matching between two successive images is shown in the inset. \subref{target-tracking-b}) Workflow of the target tracker operating at 15~Hz.}
\label{target-tracking}
\end{figure*}

This implementation of \acs{ssd}-MobileNetV2 has been trained on the standard \textsc{coco} dataset. 
The desired neural network is then trained via transfer learning on a dataset of approximately 500 images of the Jackal mobile robot from \textsf{\small Clearpath Robotics} (Fig.~\ref{target-tracking-a}). 
These images were collected from an Intel RealSense D435 camera onboard the \ac{mav} in different indoor and outdoor environments, under different lightening conditions and varying background. 
The network utilizes a $300\times300$ fixed image resizer to increase the inference speed. 
\textsf{\small Adam} optimizer was used with initial learning rate of $2\times10^{-4}$ and subsequently $10^{-4}$, $8\times10^{-5}$ and $4\times10^{-5}$ after 4500, 7000 and 10000 steps, respectively, for a total $2\times10^{4}$ steps. 
All other parameters are kept to their default values. 
The network is trained on \acs{nvidia} \textsc{gtx} 1060 \acs{gpu} and the trained model is subsequently converted into a TensorRT model to run fast inference on \acs{nvidia} Jetson-Nano \acs{gpu} onboard the \ac{mav}.

%\vspace{-0.3cm}
\subsection{Target 3D Position Estimation} \label{dnn_inplanning}

Given a bounding box obtained from the \acs{dnn}, the onboard \acs{rgbd} sensor data allow direct measurement of relative position of the target with respect to the quadrotor. 
First, all \acs{sift} features are extracted for two consecutive detection boxes and matched  using a \acs{flann}-based matcher utilizing a Lowe's ratio test~\citep{Lowe}, and a \ac{ransac} based outlier rejection was implemented. %, in a workflow showcased in  Fig.~\ref{Fig9}. 
Utilizing the disparity image from the \acs{rgbd} sensor, the $(u, v)$ position of all the inliers (i.e.\ the features on the target) in the focal plane can be converted to \acs{3d} position of the feature with respect to the camera by utilizing the disparity map. 
The average of these \acs{3d} positions provides an aggregate measurement of relative position of the target with respect to the \ac{mav}. 
This aggregate relative position measurement is used by a Kalman filter, which based on a constant-acceleration motion model for the target, returns  \acs{3d} position estimates of the target with respect to the \ac{mav}'s \acs{cog} and feeds it to the  motion planning algorithm. 
The entire target position estimation workflow is showcased in Fig.~\ref{target-tracking-b}. 
%\begin{figure}[h!] 
%	\centering
%	\includegraphics[width=0.7\linewidth]{kalman_filter.png}
%	\caption{Workflow of the target tracker operating at 15~Hz.
%		\label{Fig10}
%	}
%\end{figure} 

The effectiveness of the velocity profile prescribed in \eqref{sigmoid} is pronounced in the case of intercepting a moving target. 
Note that in case of a static navigation end goal, the first factor (the time-dependent erf) increases slowly from $0$ and converges to $1$, while the second term (distance-dependent erf) starts from $1$ and converges to $0$; 
this steers the \acs{mav} so that it starts at lower speed, progresses to near maximum allowed speeds during most of the navigation maneuver, and then smoothly slow down near its goal, thus mimicking a trapezoidal velocity profile. 
In contrast, in the case of moving target interception the distance-dependent erf in \eqref{sigmoid} converges instead to a fixed strictly positive value, allowing the \acs{mav} align its velocity with that of the target and  maintain a fixed distance to it.

\section{Implementation Results}\label{results} 

\begin{figure*}[t!]
  \includegraphics[width=\textwidth,height=4cm]{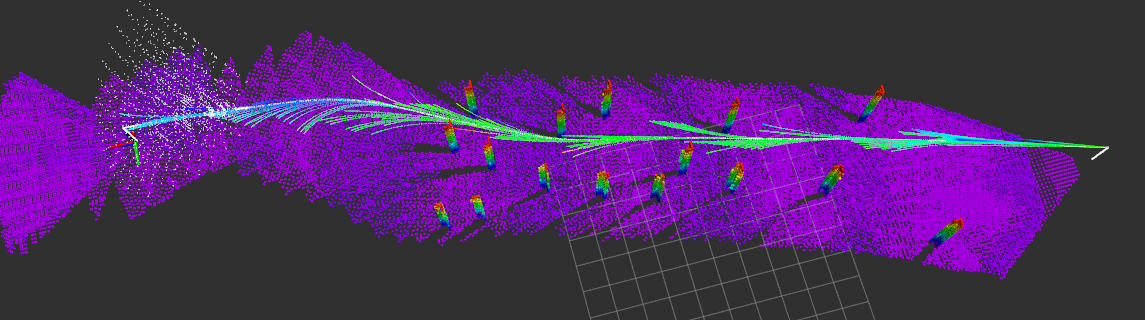}
  \caption{Receding horizon planning in a cylindrical obstacle Poisson forest. 
  \label{Fig-rhp}}
\end{figure*}

\subsection{Numerical Testing}

\begin{figure}[h!]
    \centering
	\includegraphics[width=0.8\linewidth]{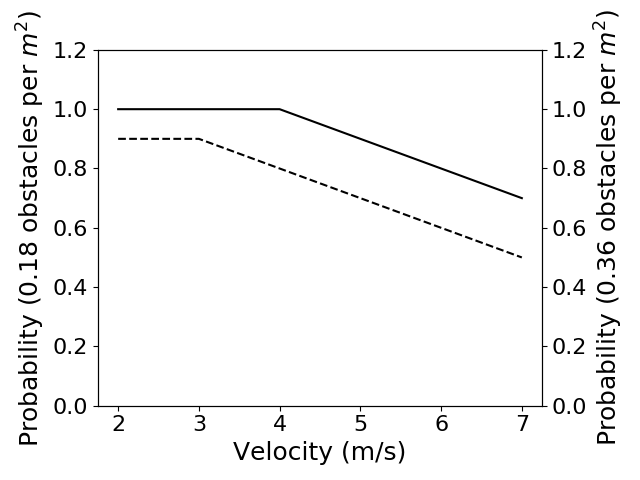} 
    \caption{Mission (target interception with collision avoidance) success  probability in  Poisson forests as a function of obstacle density and \acs{mav} speed. Solid line corresponds to left axis (0.18 obstacle/m$^2$) while dashed line corresponds to right.
    \label{maxvel-obs-density}    
    }
    %\vspace{-0.5cm}
\end{figure}

The reported planning and control framework was first tested in simulations based on the \textsc{RotorS} package~\citep{Furrer2016}, within a Poisson forest-like environment with obstacle densities of 18 and 36 obstacles in a 100 m$^2$ area.
%Fig.~\ref{Fig6b} shows the simulated environment and 
%The \acs{mav} is more likely to collide with the obstacles at higher velocities or in obstacle rich environment. 
A typical \acs{rviz} visualization of the resulting algorithm execution is shown in Fig.~\ref{Fig-rhp}. 
As the \ac{mav} flies from left to right, the figure marks in green color the different trajectories generated within each planning cycle, while the \ac{fov} grid point ensemble is shown in white toward the end of the vehicle's path.
Figure~\ref{maxvel-obs-density} shows the probability of mission success as a function of \ac{mav} speed at two different obstacle densities (cf.~\citet{SKaraman}). 

%\begin{figure}[h!]
%    \centering
%    \includegraphics[width=0.65\linewidth]{maxvel_vs_obs_density.png}
%    \caption{Mission (target interception with collision avoidance) success  probability in  Poisson forests as a function of obstacle density and \acs{mav} speed. %Adjusted to quadrotor size, these densities represent dense obstacle environments, see attached video submission. 
%    \label{Fig11}
%    }
%\end{figure}

%%%%%%%%%%%%%%%%  RADIATION-SPECIFIC %%%%%%%%%%%%%

%%%%%%%%%%%%%%%%%%%  RADIATION-SPECIFIC %%%%%%%%%%%%

\subsection{Experimental Testing}\label{ex-setup}

\begin{figure}[h!]
\centering
\includegraphics[width=0.5\textwidth]{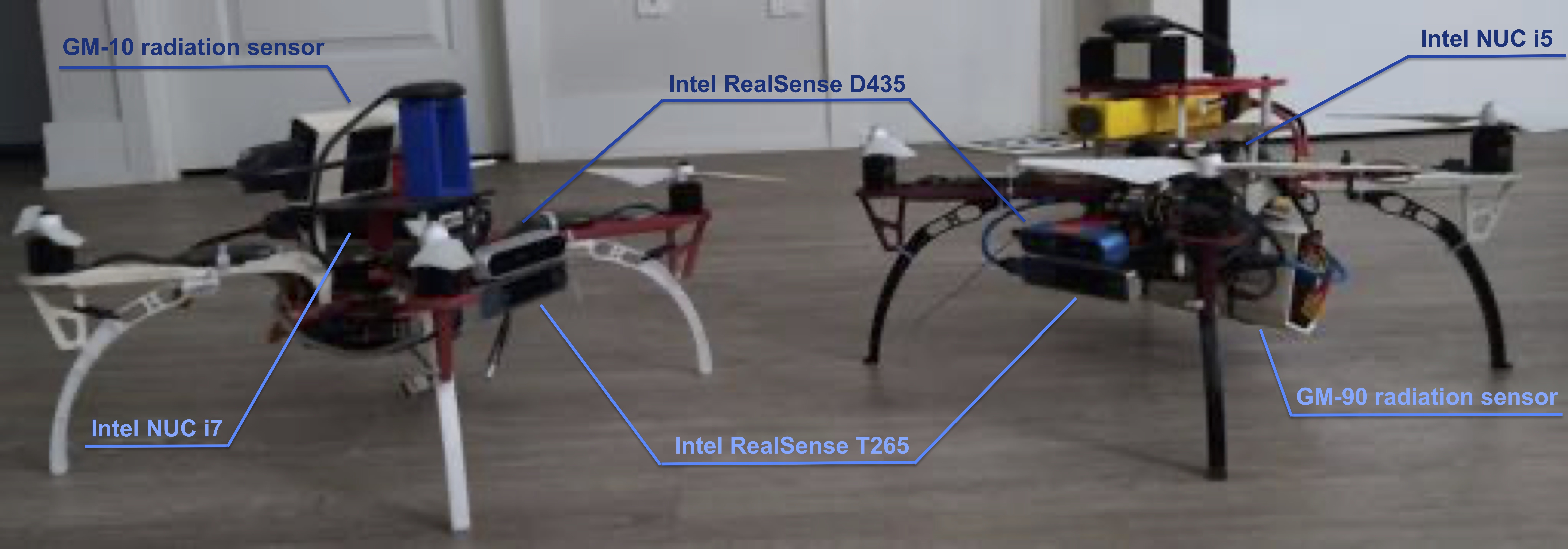}
\caption{ A pair of quadrotors fitted with \ac{cots} \ac{gm} counters.
The one on the left carries a \ac{gm}-10 counter (Blackcat Systems) , while the one on the right features a \ac{gm}-90 (more sensitive) \ac{gm} counter from the same product line.\label{figure:quads}
%-\vspace{-0.2cm}
}
\end{figure}
%and then adjusts its speed to match that of its target and maintains a fixed distance without explicit estimation of the target's velocity.

The custom-build \acp{mav} used for experimental testing shown in Fig.~\ref{figure:quads} are based on a \acs{dji} Flamewheel F450 frame.
Computationally, one of them features  an on board Intel \acs{nuc} Core i7-8650U quad core \acs{cpu}@1.9~GHz$\times$4 while the other has an Intel \acs{nuc} Core i5-7300U dual core  \acs{cpu}@2.6~GHz$\times$2. 
Both uses 16GB \acs{ram}, 128GB \acs{ssd} and a Pixhawk flight controller that is given the desired thrust magnitude and rate commands, which are then tracked using an onboard body rate controller.
A point cloud is generated by an Intel \textsf{\small RealSense}-D435 depth \acs{rgbd} camera (640$\times$480 pixel, 30~Hz) while the \textsf{\small RealSense}-T265 \acs{vi}-sensor (2 848$\times$800 pixel 30~Hz cameras with a 200~Hz \acs{imu}) is used for inertial odometry. 
This lightweight sensor package provides reliable depth information for up to 5~m. 
A voxel filter reduces the density of the generated point cloud to a uniform size of 0.125~m, which is considered adequate for typical obstacle avoidance purposes.
%\footnote{Detection of objects like cables or thin branches will require higher pointcloud densities.}
Open-\acs{vins} provides state estimation and real time odometry at 30~Hz. 
Ultimately, the pipeline's computational bottleneck is the inference script that can only be run at 15~Hz on the \acs{cpu} of the intel \acs{nuc}, and therefore the entire receding horizon planning and target tracking software is constrained at 15~Hz. 

\begin{figure}[h!]
    \centering
	\includegraphics[width=0.7\linewidth]{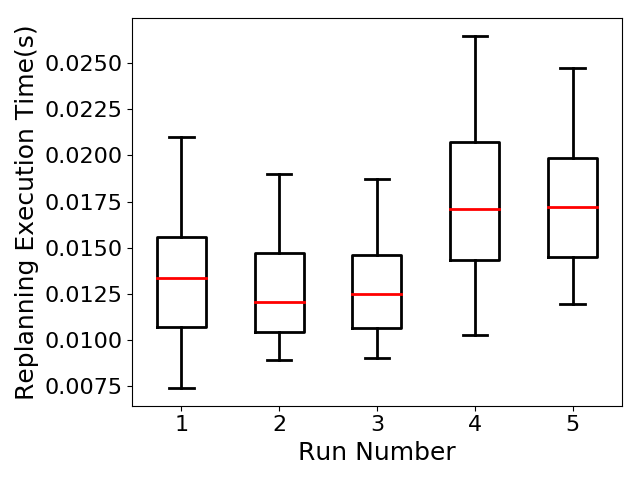} 
    \caption{Replanning execution time for the quadrotor with \acs{nuc} Core i7-8650U \acs{cpu}@1.9~GHz$\times$4 processor, averaged over five different experimental runs.
    The number of trajectories generated in each planning cycle varied between 50 to 300 while the input pointcloud size was 800-1400. Box corresponds to 5-95 quartile while the median is marked red.
    \label{Fig18} 
    }
\end{figure}

Over five different runs, each of overall trajectory length of 25 m, in both indoor and outdoor environments
%\footnote{Two of these runs are included in the paper's video supplement.} 
the 95\% quartile of execution time is well below 0.02 seconds~(Fig.~\ref{Fig18}). 
These execution times are almost half of those obtained on those systems in earlier studies~\citep{yadav_iros} that uses an optimization based trajectory generation methodology.  
In this configuration, the \acs{mav} flies safely among moderately dense outdoor obstacle environments at speeds of 4.5--5 m/s. 
%In one experiment (see video supplement) the \acs{mav} flew along a 25m trajectory around two trees between its starting and the final positions, and demonstrated collision avoidance (with the second tree) at 3.2 m/s.
%Indoor experiments demonstrated avoidance of collisions with randomly placed chairs and tables at a speed of 2.5m/s. 
These speeds surpass or are comparable to those reported in recent literature~\citep{gao,Sikang2016,Fragoso} without using any high-end and expensive sensors and their achievement is attributed to the ability to replan faster. 

%\subsection{Discussion}\label{discussion}
The video attachment that accompanies this paper features a number of experiments, the first two of which show the \ac{mav} to navigate at speeds of 2\,m/s and 4.5\,m/s in an outdoor environment for a total of 40\,m and 60\,m respectively. The video showcases first and third person views of the cluttered scene traversed by the \ac{mav},  alongside an \acs{rviz} representation of generated trajectories amongst the sensed obstacles. 
The octomap featured in the video is provided for illustration purposes only to mark obstacle locations along the path of the vehicle, and is not used for planning.  
The third experiment included in the video attachment demonstrates the \acs{mav}'s capability to avoid moving obstacles in its field of view, while the fourth showcases receding horizon navigation with combined obstacle and target tracking abilities.
The latter utilizes a neural network with feature matching between two subsequent images.

Ultimately, the \acs{mav}'s speed will be limited primarily by the computational capabilities and the need to implement a safe emergency stop in the absence of an effective motion plan---a possibility which cannot be eliminated in purely reactive and local planning methods.
Since purely reactive approaches are supposed to rely \emph{exclusively} on local information, convergence to the navigation goal cannot be guaranteed for all possible scenarios. 
(The reported reactive planning architecture can however be integrated with an exact global planner, and overcome these limitations at the price of using global information~\citep{yadav_CDC2021}.)
%Along with walls and high obstacle density instances, the robot may also get trapped into concave obstacles.
%The primary focus of the this particular work is \emph{push the boundaries of what can be achieved in purely reactive but deliberate quadrotor navigation in cluttered spaces.}
This investigation naturally exposes the limits of purely reactive motion planning approaches.
It is expected that knowledge of those limits, can guide the development of hybrid (local \& global~\citep{Yadav}) \ac{mav} motion planning methodologies destined for deployment in environments where uncertainty is reasonably well characterized, in order to complement each other and operate robustly in real-world scenarios.

While on one hand the of perception from trajectory generation can induce latencies, their co-design using powerfull on board computation and the incorporation of pre-trained neural networks for trajectory generation can further boost vehicle speeds to impressive levels~\citep{aLoquercio}.
To reach speeds up to 10 m/s, however, one would also require a very favorable power-to-weight ratio (e.g.  $\sim$4.0~\citep{aLoquercio}), which may be challenging to maintain depending on the \emph{mission-mandated sensor payload}.
In this paper, the \acp{mav} featured a power-to-weight ratio in the 1.55--1.75 range.
Besides an effort to achieve a more favorable power-to-weight ratio, we postulate that additional speed improvements can be achieved with the use of event cameras~\citep{david_falanga}.

\subsection{Radiation Detection}

The radiation detection algorithm is based on a Neyman-Pearson based, fixed time interval binary hypothesis test (refer to \cite{PSPT_Automatica} for a more detailed exposition).
At the heart of the test is a likelihood ratio of a statistic $L_T$~\eqref{Eq:LT} calculated based on the history of relative distance between the airborne radiation sensor and the hypothesized source in addition to the aggregated counts over the sensor's (predetermined) integration time interval $T$. This likelihood ratio is compared against a fixed threshold value $\gamma$ that also depends on the relative distance and the acceptable bound on the probability of false alarm $P_{\mathsf{FA}}$. The optimal value of $p^{*}$ is obtained by solving~\eqref{Eq:PF} and then the threshold is calculated by evaluating~\eqref{LambdaPrime} at $p^{*}$.
The remaining parameters for this test, including background naturally occurring (background) radiation rate and sensor characteristics, are determined by computational and experimental calibration processes.

%Due to exposure risks involved in conducting experiments with gamma ray sources, the complete radiation detection assessment contains two parts.
Our detection tests involve a sequence of controlled experiments in which the efficiency of the radiation sensors and the variation of its expected count rate as a function of its distance to the source is estimated.
%, while the computational portion involves a Monte Carlo simulation for the estimation of the minimum length $T$ of the time interval for radiation measurement collection that is more likely to yield a confident and accurate classification of the target relative to its radioactivity.
The \acp{mav} featured in Fig.~\ref{figure:quads} were deployed in both indoor and outdoor experiments, where their task was to locate and intercept a ground target (the remotely controlled \textsf{\small ClearPath Robotics Jackal} in Fig.~\ref{Fig1}) moving along an unspecified path with unknown but bounded speed. 
The ground robot carried an approximately 8~$\mu$Ci radioactivity source which the \acp{mav} had to detect.
%The variation in the distance between sensor and source can also be attributed to the motion variations of the target's path and speed as the human operator was attempting to avoid interception.

Each experiment involved the \acs{mav} tracking the ground vehicle for certain time $T$. 
To minimize radiation exposure risk,  Monte Carlo simulations using \acs{gazebo} have been performed to compliment the experimental validation. 
The counts were generated using thinning algorithm in the simulations~\citep{Pasupathy}.  These simulations were used for the estimation of the minimum length $T$ of the time interval for radiation measurement collection that is more likely to yield a confident and accurate classification of the target relative to its radioactivity.
This overall process suggested a sensor integration window set at $T = 100$ seconds for the \textsc{gm}-10 counter (at the median of the distribution with 5\% and 95\% percentiles estimated at 71 and 136 seconds, respectively), for a radiation source of activity around $8.2\,\mu$Ci.
During that fixed time interval, the distance between sensor and source should be at most within 3--2.5 m, otherwise the source emissions completely blend into the background, rendering detection extremely unlikely given the \ac{mav}'s operating time constraints imposed by on-board power.
For that same source and range to sensor, the \textsc{gm}-90 counter appeared to need approximately $T = 70$ seconds, with a 5\% and 95\% percentiles at 65 and 96 seconds, respectively.

\begin{figure}[h!]
    \centering
	\includegraphics[width=0.8\linewidth]
	{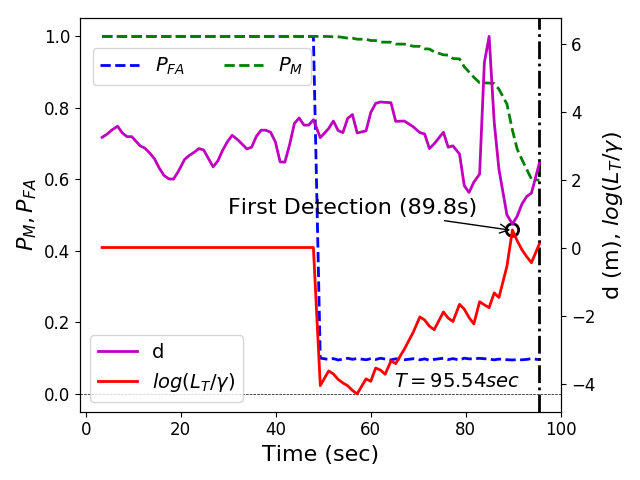} 
    \caption{Detection parameters for GM-10 sensor as a function of decision time $T$. Bound on probability of false alarm: dashed blue; bound on the probability of missed detection: dashed green; ratio $\log\sfrac{L_T}{\gamma}$: solid red; sensor-source distance: solid magenta.
    \label{figure:detection_exp}    
    }
\end{figure}

The receding horizon planning and control strategy of Section~\ref{section:RHP} ensures that the \ac{mav} closes this distance as fast as possible, thus enabling the onboard Geiger counters to collect informative measurements.
%In lieu of any prior information about the path and speed of the target vehicle and practical limitations of the methodology discussed in section~\ref{targettracking}, the relative distance cannot be arbitrarily reduced to zero. 
The \acs{mav} has to maintain a certain minimum distance from the target to keep it in its camera's limited \acs{fov}; as a result, the relative distance should not be  arbitrarily reduced. 
%The maximum achievable relative distance too is restricted by the ability of the neural network to detect small objects and  
Variations in the relative distance between sensor and source can be attributed to motion perturbations, as the (remotely operated) target performs avoidance maneuvers.
Although generally robust, the  neural network does not furnish  guarantees against false target identification, and this can contribute to relative distance estimate outliers (see e.g.\  Fig.~\ref{figure:detection_exp}, around the 85\textsuperscript{th} second).
%In lieu of all this, the relative distance reported in Figs.~\ref{figure:detection_exp}--\ref{figure:GM90} has minor variations. 

%Earlier work~\cite{PSPT_Automatica,Sun} has established that this motion strategy is optimal for minimizing (a Chernoff bound on) the probability of missed detection $P_\mathsf{M}$.
%\textcolor{red}{
%In lieu of any prior information about the path and speed of the target vehicle, \eqref{mav-control}--\eqref{minsnap} can only guarantee uniform ultimate boundedness of the errors \eqref{errors}, a fact that is reflected in Figs.~\ref{figure:detection_exp}--\ref{figure:GM90}.
%}

Figure~\ref{figure:detection_exp} presents the results of one radiation detection experiment conducted in an indoor environment (an abandoned warehouse) using the \ac{mav} that has GM-10 counter.
It shows the evolution of the estimate of the relative distance $d$ between the \ac{mav} and the ground robot as the latter moves with unknown and time-varying speed.  
The relative distance is estimated in real time via the target tracking pipeline described in the section on target tracking.
The dashed curves in Fig.~\ref{figure:detection_exp} indicate the evolution of Chernoff bounds on the probability of false alarm, $P_\mathsf{FA}$, and probability of missed detection $P_\mathsf{M}$~\citep{PSPT_Automatica}.
The bound on the probability of false alarm appears to drop below the acceptable upper limit after approximately 50 seconds from the start of the experimental run, after which the bound on the probability of missed detection $P_\mathsf{M}$ also starts to slowly decrease monotonically---the latter is a decreasing function of the sensor integration time and distance between sensor and source~\citep{PSPT_Automatica}.
The graph of the logarithm of the likelihood ratio $L_T$ over the detection threshold $\gamma$ over time is marked in red; this process is stochastic because it depends directly on the arrival time of gamma rays on the sensor.
The initial segment of the red curve corresponds to the initial time period during which the constraint on $P_{FA}$ has not been satisfied and $\log\sfrac{L_T}{\gamma}$ has been kept at 0.
The experiment is concluded at $95.54$ seconds and the likelihood ratio $L_T$ exceeds its threshold value at $89.8$ seconds indicating the presence of the radiation source on the ground target (marked with a black circle in the plot). 
The likelihood ratio had actually crossed the threshold before that time, but the experiment was continued because that event was  observed significantly earlier than the recommended sensor integration window.

\begin{figure*}[t]
\centering

\subfloat[GM90 Indoor Run]{%
  \includegraphics[width=0.4\textwidth]{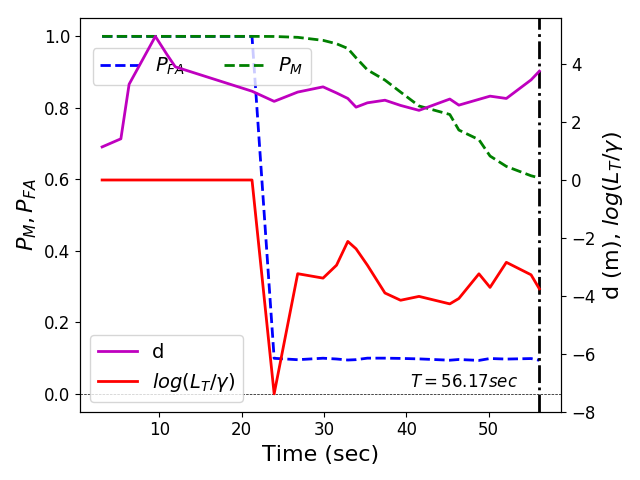}%
  \label{figure:GM90indoors}
}
\subfloat[GM90 Outdoor Run]{%
    \centering
  \includegraphics[width=0.4\textwidth]{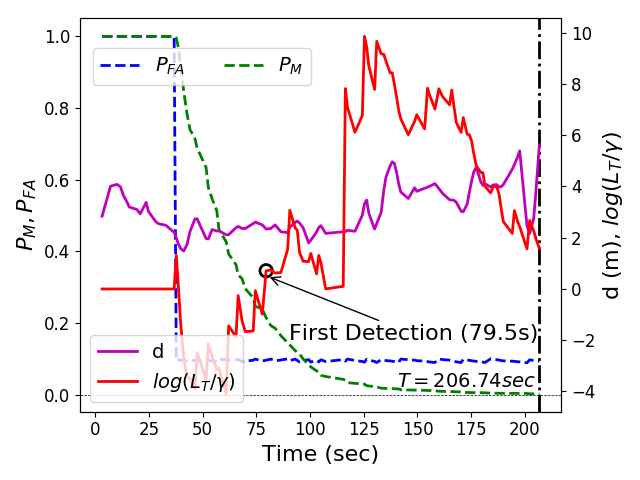}%
  \label{figure:GM90outdoors}
}
\caption{Detection parameters for GM-90 sensor as a function of decision time $T$. Bound on probability of false alarm: dashed blue; bound on the probability of missed detection: dashed green; ratio $\log\sfrac{L_T}{\gamma}$: solid red; sensor-source distance: solid magenta. \subref{figure:GM90indoors}) Detection time interval $56$ seconds. \subref{figure:GM90outdoors}) Detection time interval $206$ seconds.}
\label{figure:GM90}
%}
\end{figure*}

Figures~\ref{figure:GM90indoors} and \ref{figure:GM90outdoors} showcase two different runs where the \acs{mav} featuring the \textsc{gm}-90 counter was utilized.
The experimental run of Fig.~\ref{figure:GM90indoors} shows an instance where the \ac{mav} did not have enough time to detect the source.
This experiment was performed in the same indoor facility as that used for the run of Fig.~\ref{figure:detection_exp}.
Here, the radiation sensor integration window is 56 seconds.
The bound on the probability of missed detection is then still around 0.6, comparable to the conditions under which the detection of Fig.~\ref{figure:detection_exp} was achieved, but this $T$ is below the 5\% percentile for the recommended exposure time.

%\begin{figure}[h!]
%    \centering
%	\includegraphics[width=0.75\linewidth]
%	{detection_exp2_gm90.png} 
%    \caption{Detection parameters as a function of decision time $T$. Bound on probability of false alarm: dashed blue; bound on the probability of missed detection: dashed green; ratio $\log\sfrac{L_T}{\gamma}$: solid red; sensor-source distance: solid magenta. Note that the evolution of the likelihood ratio $\log \sfrac{L_T}{\gamma}$ is measured on the right vertical axis.
%    \label{figure:GM90indoors}    
%    }
%\end{figure}

%\begin{figure}[h!]
%    \centering
%	\includegraphics[width=0.75\linewidth]
%	{detection_exp1_gm90.png} 
%    \caption{Detection parameters as a function of decision time $T$. Bound on probability of false alarm: dashed blue; bound on the probability of missed detection: dashed green; ratio $\log\sfrac{L_T}{\gamma}$: solid red, measured on the right axis.
%    \label{figure:GM90outdoors}  
%    \vspace{-0.5cm}
%    }
%\end{figure}
Figure~\ref{figure:GM90outdoors} depicts the results of a longer chase by the \ac{mav} carrying the \textsc{gm}-90 counter conducted outdoors.
This time, the integration window was extended to more than 200 seconds.
In addition to the effect of sensor integration window length on detection accuracy, Fig.~\ref{figure:GM90outdoors} shows more clearly the evolution of the bounds on the decision test's error probabilities $P_\mathsf{FA}$ and $P_\mathsf{M}$.
At the time of decision, the bound on the probability of miss, $P_\mathsf{M}$ is almost zero, indicating very high probability for accurate decision-making.
Although the statistic $\log\sfrac{L_T}{\gamma}$ crosses becomes positive for the first time shortly after 70 seconds, at that time the bound $P_\mathsf{M}$ is around 0.3.
It is of interest that towards the end of the integration window, the statistic $\log\sfrac{L_T}{\gamma}$ decreases, most likely due to the target being able to open up its distance with respect to the pursuing \ac{mav}---which by that time was experiencing a drop in its power reserves; same trend can be noticed in Fig.~\ref{figure:GM90indoors}.

The accompanying video attachment includes two experiments  of target tracking with radiation detection (experiments \#5 and \#6).
These are cases where the Jackal robot is steered manually while carrying an $8.2\,\mu$Ci source.
The plot at the bottom left of the video represents graphically the evolution of the detection parameters as depicted in Figs.~\ref{figure:detection_exp}~and~\ref{figure:GM90outdoors}. To reduce the size of the video file, only selected portions of these experimental runs are included and the video is accelerated at four times the normal speed.

%\vspace{-0.3cm}
\section{Conclusions}\label{conclusions}
The challenges that a completely autonomous \ac{mav} equipped with short-ranged sensors faces when tasked to navigate in a completely unknown and cluttered environment are more pronounced when the complete data and signal processing (such as radiation detection) pipeline needs to be run onboard, and the motion of the vehicle can have adverse effect on the quality and quantity of the data.
Under these conditions, a motion planner that aims at operating the robot within a reasonable safety envelope has to strike a balance between safety,  platform limitations, and   mission-informed active sensing.
In this context, an adaptive, purely reactive receding horizon motion planning and control strategy has been developed that co-designs the planning, safe navigation, target tracking and decision-making components.
%within the spirit of receding horizon control appear to be appropriate and effective, and this paper reports on one such method with improved and novel characteristics. 
Not only can such a navigation strategy be remarkably effective even in the absence of global environment and platform information precludes formal completeness guarantees, but it can  also can be integrated with an exact global planner when prior knowledge permits, to furnish formal performance guarantees.
The work thus pushes the envelope on what is achievable with \emph{purely reactive} but deliberate \ac{mav} navigation in cluttered environments, particularly at the low-end of the technology and sensor sophistication spectrum.

\begin{acks}
Special thanks to Paul Huang and his RPNG group, specifically to Kevin Eckenhoff and Patrick Geneva for his contributions on perception.
\end{acks}
\begin{funding}
This work was made possible in part by DTRA grant \#HDTRA1-16-1-0039 and ARL grant \#W911NF-20-2-0098.
\end{funding}

\balance
\bibliographystyle{SageH}      
%\bibliography{Effect_of_noise_on_radiation_detection_R0,Networking,DecisionMaking,decision_graph} 
\bibliography{Networking,DecisionMaking,decision_graph} 

\end{document}